\documentclass[10pt,twocolumn,letterpaper]{article}

\usepackage[pagenumbers]{cvpr} 
\usepackage{caption}
\usepackage{booktabs}
\usepackage{tabularx}
\usepackage{pifont}
\usepackage{graphicx}
\usepackage{array}
\usepackage{xcolor}
\usepackage{multirow}
\usepackage{lineno}
\usepackage[normalem]{ulem}

\newcommand{\cmark}{\textcolor{green!55!black}{\ding{51}}} 
\newcommand{\xmark}{\textcolor{red!75!black}{\ding{55}}} 
\newcolumntype{Y}{>{\centering\arraybackslash}X}

\usepackage{caption}

\definecolor{cvprblue}{rgb}{0.21,0.49,0.74}
\usepackage[pagebackref,breaklinks,colorlinks,allcolors=cvprblue]{hyperref}

\title{VGGT-$\mathbf{\forall}$ : 3D Scene Affordance Discovery from Images Alone}

\author{%
  Patrick Kwon \\
  Institute of Artificial Intelligence \\
  University of Central Florida\\
  Orlando, FL 32817 \\
  \texttt{yo564250@ucf.edu} \\
  \and
  Chen Chen \\
  Institute of Artificial Intelligence \\
  University of Central Florida\\
  Orlando, FL 32817 \\
  \texttt{chen.chen@ucf.edu} \\
}

\newcommand{\fullscene}[1]{%
    \includegraphics[width=\linewidth]{#1}%
}
\newcommand{\instanceimg}[1]{%
    \includegraphics[width=\linewidth]{#1}%
}
\newcommand{\qualScene}[8]{%
    \noindent\textbf{#1}\par\vspace{1mm}
    \noindent
    \begin{minipage}[t]{0.49\textwidth}
        \centering
        \fullscene{#2}\\[-1pt]
        {\small Ground Truth Point Cloud \& VGGT-$\forall^\dagger$ predicted affordances}
    \end{minipage}%
    \hfill%
    \begin{minipage}[t]{0.49\textwidth}
        \centering
        \fullscene{#3}\\[-1pt]
        {\small VGGT-$\forall$ Point Cloud \& VGGT-$\forall$ predicted affordances}
    \end{minipage}

    \par\vspace{2mm}

    \noindent
    \begin{minipage}[t]{0.195\textwidth}
        \centering
        \instanceimg{#4}\\[-1pt]
        {\small \textbf{Origin}}
    \end{minipage}%
    \hfill%
    \begin{minipage}[t]{0.195\textwidth}
        \centering
        \instanceimg{#5}\\[-1pt]
        {\small \textbf{VGGT-$\forall^\dagger$}}
    \end{minipage}%
    \hfill%
    \begin{minipage}[t]{0.195\textwidth}
        \centering
        \instanceimg{#6}\\[-1pt]
        {\small Volt}
    \end{minipage}%
    \hfill%
    \begin{minipage}[t]{0.195\textwidth}
        \centering
        \instanceimg{#7}\\[-1pt]
        {\small OVA-Fields}
    \end{minipage}%
    \hfill%
    \begin{minipage}[t]{0.195\textwidth}
        \centering
        \instanceimg{#8}\\[-1pt]
        {\small LERF}
    \end{minipage}
    \par
}

\begin{document}

\twocolumn[{%
    \renewcommand\twocolumn[1][]{#1}%
    \maketitle
    
    \begin{center}
        \centering
        \vspace{-12pt}
        \includegraphics[width=1.0\textwidth]{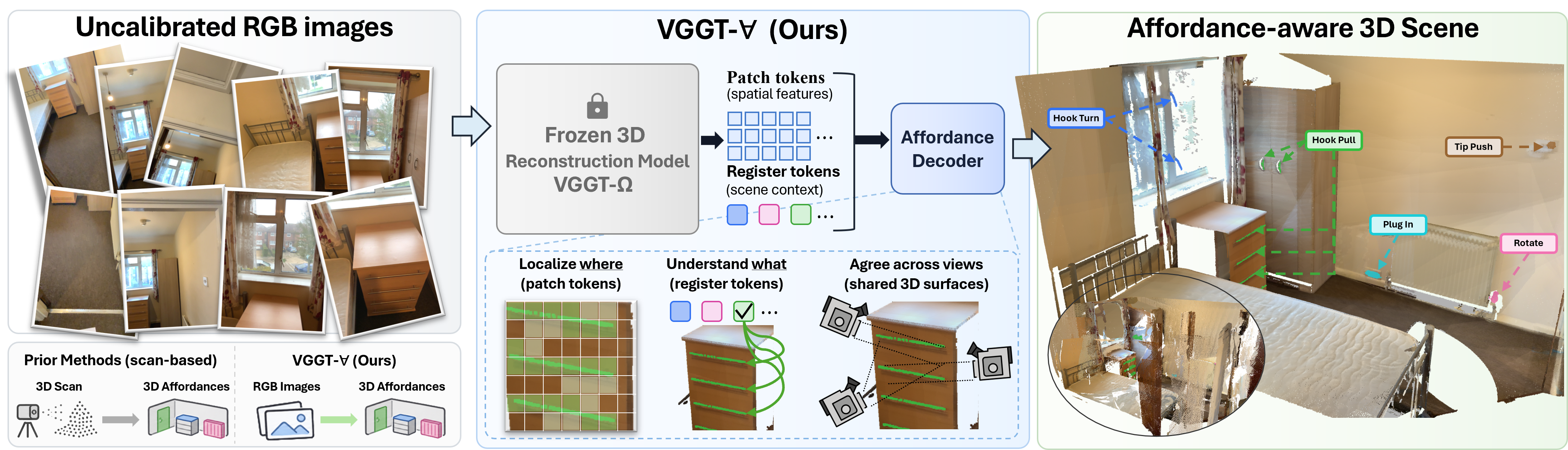}
        \captionof{figure}{\textbf{Image-based scene affordance discovery.} Unlike prior methods that require high quality 3D scans, VGGT-$\forall$ jointly performs 3D reconstruction and affordance discovery from uncalibrated RGB images, enabling accessible scene-level affordance discovery.}
        \label{fig:title_figure}
    \end{center}%
}]

\begin{abstract}
Scene affordances describe where and how an environment can be interacted with, yet existing methods typically rely on high-fidelity 3D scans. To solve this, we introduce \textbf{VGGT-$\forall$}, a feed-forward framework that jointly reconstructs 3D scene geometry and discovers affordances from RGB images alone. Our key insight is that representations from reconstruction provide preliminary spatial and semantic cues that can be developed into affordance predictions. We provide an affordance decoder that combines spatial localization with semantic reasoning, using surface-level supervision and cross-view consistency to recover coherent instances. For scan-based evaluation, we introduce \textbf{VGGT-$\forall^\dagger$} that uses VGGT-$\forall$ predictions to guide localized segmentation. Experiments demonstrate that VGGT-$\forall$ outperforms adapted baselines on image-reconstructed geometry, while VGGT-$\forall^{\dagger}$ surpasses the strongest published SceneFun3D baseline while processing less than 20\% of scan points, highlighting both accessible 3D affordance discovery and efficient scan-assisted refinement. Project Page: \url{https://yj7082126.github.io/vggt_forall/}
\end{abstract}

    
\section{Introduction}
\label{sec:intro}

Gibson introduced \emph{affordance} to describe the opportunities for interaction that an environment offers; for instance, a door handle affords pulling, while a power outlet affords plugging~\cite{gibson1979ecological}. In visual understanding, affordance goes beyond mere object identification by describing how it can be interacted with, and which functional regions support those actions~\cite{do2018affordancenet,deng2021affordancenet,yang2023iag}. Such understanding connects perception with action, and has been explored for robotic manipulation~\cite{bahl2023affordances,xu2021Affkpnet,wu2023environment}, imitation and demonstration learning~\cite{mandlekar2022matters,bahl2023affordances}, human-robot interaction~\cite{braud2020robot}, visual navigation~\cite{nagarajan2020affordance}, and augmented or virtual reality~\cite{raikwar2019cubevr}. 

Existing methods for scene-level affordances however, require inputs beyond ordinary images. 3D instance segmentation methods such as Volt~\cite{yilmaz2026volt} directly segment high fidelity scene point clouds, which require scanning equipment that limits practical accessibility. Purely image-based methods mostly focus on objects~\cite{do2018affordancenet, wu2026affordany}, and while methods like OVA-Fields~\cite{su2025ovafields} use 2D segmentations that are integrated into 3D scenes, they still depend on external 3D geometric inputs and target-specific queries. One alternative is to use scan-based segmentation models to reconstructed point clouds; however, the geometric distribution shift from the scans used for training undermines their reliability~\cite{qu2026segvggt}.


To address these limitations, we introduce \textbf{VGGT-$\forall$}, where $\forall$ (``for all'') reflects accessible affordance discovery. Our feed-forward framework jointly reconstructs scene geometry and discovers 3D affordances from uncalibrated images alone. Unlike SegVGGT~\cite{qu2026segvggt}, which integrates object queries into VGGT~\cite{wang2025vggt} for object discovery, we leverage the complementary patch and scene-register representations of frozen VGGT-$\Omega$~\cite{wang2026vggtomega} for affordance discovery. Our insight is that these tokens provide complementary spatial and semantic cues: patch tokens indicate \emph{where} affordances may occur, while register tokens provide context about \emph{what} affordances may be present.

\begin{table}[t]
    \centering
    \caption{
        Comparison with prior affordance methods.
        VGGT-$\forall$ enables scene-level 3D affordance
        discovery from images alone.
    }
    \label{tab:method_comparison}

    \renewcommand{\arraystretch}{1.12}
    \setlength{\tabcolsep}{3pt}

    \begin{tabular}{@{}lcccc@{}}
        \toprule
        \textbf{Method} &
        \textbf{\shortstack{3D\\Input}} &
        \textbf{\shortstack{Scene-\\Level}} &
        \textbf{\shortstack{RGB-\\Only}} &
        \textbf{\shortstack{Query-\\Free}} \\
        \midrule

        AffordanceNet~\cite{do2018affordancenet}
            & \xmark & \xmark & \cmark & \cmark \\

        3D AffordanceNet~\cite{deng2021affordancenet}
            & \cmark & \xmark & \xmark & \cmark \\

        IAGNet~\cite{yang2023iag}
            & \cmark & \xmark & \xmark & \xmark \\

        Affostruction~\cite{park2026affostruction}
            & \cmark & \xmark & \xmark & \xmark \\

        \midrule

        Mask3D-F~\cite{delitzas2024scenefun3d,schult2023mask3d}
            & \cmark & \cmark & \xmark & \cmark \\

        Fun3DU~\cite{corsetti2025fun3du}
            & \cmark & \cmark & \xmark & \xmark \\

        OVA-Fields~\cite{su2025ovafields}
            & \cmark & \cmark & \xmark & \xmark \\

        AffordMatcher~\cite{vu2026affordmatcher}
            & \cmark & \cmark & \xmark & \xmark \\

        Volt~\cite{yilmaz2026volt}
            & \cmark & \cmark & \xmark & \cmark \\

        \midrule

        \textbf{VGGT-$\forall$ (Ours)}
            & \xmark & \cmark & \cmark & \cmark \\

        \bottomrule
    \end{tabular}

    \vspace{2pt}
    \begin{minipage}{\columnwidth}
        \scriptsize
        \textit{3D Input}: requires externally supplied scans,
        point clouds, or depth measurements.
        \textit{Scene-Level}: predicts 3D scene affordances.
        \textit{RGB-Only}: operates on uncalibrated RGB images
        without external 3D input.
        \textit{Query-Free}: requires no target-specific
        language or visual interaction prompt.
    \end{minipage}
    \vspace{-16pt}
\end{table}

Building upon a frozen VGGT-$\Omega$~\cite{wang2026vggtomega}, we introduce a dense affordance decoder that combines multiscale spatial features with register-conditioned class queries. \uline{To our knowledge, this is the first work to explicitly leverage VGGT-$\Omega$ register (a.k.a scene) tokens for scene affordance.} During training, surface-level supervision and cross-view consistency encourage coherent predictions across observations of the same physical surface. At inference, predicted masks are lifted onto reconstructed geometry and associated across views to form persistent 3D affordance instances, discovering all supported affordances without externally supplied point clouds, depth maps, or camera poses.

Established benchmarks~\cite{delitzas2024scenefun3d} evaluate affordances on native point clouds. Therefore, to enable direct and fair comparisons, we also introduce \textbf{VGGT-$\forall^\dagger$}, a scan-assisted variant that uses VGGT-$\forall$'s discovered affordance regions to guide localized 3D segmentation. A crop-adapted segmentation network aligns these affordances with native point clouds, recovering finer boundaries and filtering false detections without requiring full-scene segmentation inference.

We evaluate VGGT-$\forall$ on SceneFun3D~\cite{delitzas2024scenefun3d} under both RGB-only and native point-cloud settings. Under our RGB-only evaluation protocol, VGGT-$\forall$ achieves 38.79 $\mathrm{AP}_{50}$ on reconstructed validation surfaces, substantially outperforming an adapted SegVGGT baseline. With native point clouds, VGGT-$\forall^\dagger$ further achieves 40.4 mAP on validation and 23.0 mAP on the hidden test set, surpassing Volt~\cite{yilmaz2026volt}, the strongest peer-reviewed baseline reported on the SceneFun3D Task 1 leaderboard. These results demonstrate the effectiveness of discovering affordances directly from image without pre-existing scene geometry, while establishing competitive performance under an established, independently evaluated scan-based benchmark.

Our main contributions are summarized as follows:
\begin{itemize}
    
    
    \item We introduce VGGT-$\forall$, the first framework to achieve scene-level 3D affordance discovery from uncalibrated RGB images alone, jointly reconstructing scenes with affordances without external 3D data at inference.
    \item We repurpose a frozen VGGT-$\Omega$ for affordance discovery, leveraging its latent spatial and semantic cues that are relevant to understanding where and how interactions will occur, developing them into 3D affordance instances.
    \item We introduce \textbf{VGGT-$\forall^\dagger$}, an extension that aligns VGGT-$\forall$ output affordances to the input point cloud, outperforming Volt on SceneFun3D while processing only 18.74$\%$ of the scanned points and accelerating the segmentation stage up to 3.6$\times$ with reduced GPU memory.
\end{itemize}












\section{Related Work}
\label{sec:related}

\subsection{Affordance Discovery}
\label{sec:related_affordance}

\paragraph{Object-level affordance.} 
Early affordance understanding focused on identifying interaction regions of individual objects, with AffordanceNet~\cite{do2018affordancenet} introducing end-to-end pixel-level affordance detection and 3D AffordanceNet~\cite{deng2021affordancenet} extending this to 3D object surfaces. Subsequent work focus on richer sources of functional cues, such as part transfer~\cite{li2023locate}, human--object interactions~\cite{yang2023iag}, and spatial interaction priors beyond contact~\cite{kim2024ComA}. More recent works also support language guidance and open-vocabulary grounding~\cite{li2024LASO,tuan2024OpenAD}, with stronger generalization through cross-modal representations~\cite{gao2025MIFAG,lu2025GEAL,li2025Glance}. More recently, AffordAny~\cite{wu2026affordany} enabled instruction-conditioned 3D part grounding through monocular object reconstruction. Despite these advances, existing methods focus on identifying affordances within individual objects or parts, rather than discovering affordances across an entire 3D scene. Extending affordance to scenes requires locating diverse functional regions across multiple objects and scene structures, a challenge addressed by VGGT-$\forall$.

\begin{figure*}
    \centering
    \includegraphics[width=1\textwidth]{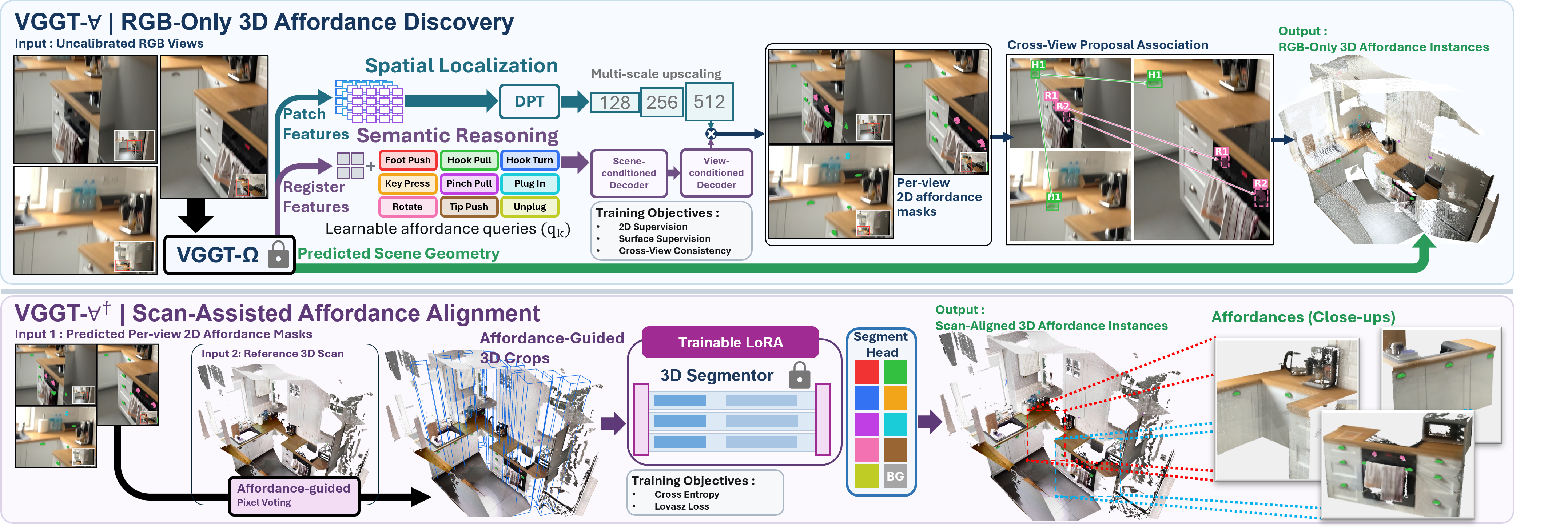}
    \caption{\textbf{Overall Architecture.} (top) VGGT-$\forall$ reconstructs 3D geometry and discovers affordances from images alone. Patch and register tokens from VGGT-$\Omega$ are processed through the spatial and semantic pathways of our decoder to predict per-view affordance masks. Through proposal association, 3D affordance instances are recovered on the reconstructed scene. (bottom) When a point cloud is required, VGGT-$\forall^\dagger$ reuses these per-view masks to identify 3D crops over the point cloud, and uses a LoRA-adapted segmentation model to produce scan-aligned affordance instances. Best viewed when zoomed.}
    \label{fig:main_figure}
    \vspace{-12pt}
\end{figure*}

\paragraph{Scene-level affordance.} Scene-level affordance understanding extends affordance to complete environments through interaction mapping~\cite{li2019putting,nagarajan2020affordance,murlabadia2023multilabel} and fine-grained functionality prediction~\cite{delitzas2024scenefun3d}. One line of approach extends 3D instance segmentation networks for affordances, building on architectures such as PointGroup~\cite{jiang2020pointgroup}, SoftGroup~\cite{vu2022softgroup}, and Mask3D~\cite{schult2023mask3d}. Recent models such as Volt~\cite{yilmaz2026volt} further improve performance through a data-efficient training recipe. While effective, these methods rely on detailed point clouds, which have a different distribution from reconstructed scenes, limiting their direct applicability and motivates discovering affordances from uncalibrated images rather than 3D scans.

Other approaches incorporate additional interaction cues for affordance discovery. Ego-SAG~\cite{liu2024egosag} and FunRec~\cite{delitzas2026funrec} leverage egocentric interactions, while Fun3DU~\cite{corsetti2025fun3du} and OVA-Fields~\cite{su2025ovafields} exploit visual features and language guidance to identify functional regions in 3D scenes. TASA~\cite{he2026tasa} further combines task-aware 2D predictions with local geometric refinement, and AffordMatcher~\cite{vu2026affordmatcher} establishes correspondences between visual signifiers and 3D scene instances. As summarized in Table~\ref{tab:method_comparison}, these method effectively utilize semantic cues for affordance discovery, but require 3D scene information and often assume prior knowledge of the scene's functionality, which may be unavailable in practical applications. In contrast, VGGT-$\forall$ directly discovers scene-level 3D affordances from uncalibrated images without external geometry or target queries.

\subsection{Feed-Forward 3D Reconstruction}
\label{sec:related_vggt} 
Traditional 3D reconstruction relies on geometric correspondence and multi-stage optimization~\cite{schonberger2016sfm}, while neural representations such as NeRF and 3D Gaussian Splatting commonly require scene-specific optimization~\cite{mildenhall2020nerf,kerbl20233d}. Learning-based approaches progressively introduced transferable geometric priors, from cross-view representation learning~\cite{weinzaepfel2022croco,weinzaepfel2023crocov2} to pairwise reconstruction and matching~\cite{wang2024dust3r,leroy2024mast3r}, followed by scalable multi-view reconstruction~\cite{wang2025spann3r,wang2025cut3r,yang2025fast3r}. VGGT~\cite{wang2025vggt} further unifies cameras, depth, point maps, and correspondences within a single feed-forward transformer, while VGGT-$\Omega$~\cite{wang2026vggtomega} introduces register attention and identifies its register tokens as compact carriers of global scene information. 

Recent works such as PanSt3R~\cite{zust2025panst3r} and IGGT~\cite{li2026iggt} extend feed-forward 3D reconstruction toward scene understanding. SegVGGT~\cite{qu2026segvggt} further integrates object queries into VGGT for joint reconstruction and 3D instance segmentation. Unlike such object-centric approaches, we focus on a scene-level approach and repurpose both spatial tokens for localization and scene-level register tokens for semantic context, extending feed-forward reconstruction toward functional scene understanding.

\section{Methods}
\label{sec:methods}

Our main objective is to jointly recover the 3D geometric and functional structure of a scene from uncalibrated images. Given $N$ uncalibrated RGB images $\{I_i\}_{i=1}^{N}$, where $I_i \in \mathbb{R}^{3\times H\times W}$, VGGT-$\forall$ predicts scene geometry $\mathcal{G} = \{\mathbf{x}_m\}_{m=1}^{M}$, where $\mathbf{x}_m \in \mathbb{R}^{3}$, and a set of $N_A$ affordance instances across $K$ categories:
\begin{equation}
\mathcal A={(\mathcal I_j,k_j,s_j)}_{j=1}^{N_A}
\label{eq}
\end{equation}
where $\mathcal{I}_j \subseteq \{1,\ldots,M\}$ indexes the reconstructed points belonging to instance $j$, $k_j \in \{1,\ldots,K\}$ denotes its affordance category\footnote{The categories are foot push, hook pull, hook turn, key press, pinch pull, plug in, rotate, tip push, and unplug. See Supplementary for details.}, and $s_j \in [0,1]$ represents its confidence. Thus, $\mathcal I_j$ identifies \emph{where} an interaction is possible, while $k_j$ specifies \emph{what} interaction the region affords. 



\subsection{Preliminaries}
\label{sec:prelim}

\paragraph{VGGT.}
Given $N$ images, VGGT~\cite{wang2025vggt} performs feed-forward reconstruction by jointly predicting camera parameters, depth maps, and point maps. Each image is encoded into spatial patch tokens $\mathbf{z}_i^F$, which are processed by alternating frame-level and inter-frame attention to capture 3D geometric information. VGGT-$\Omega$~\cite{wang2026vggtomega} introduces dedicated register attention for register tokens $\mathbf{Z}_i^R$ to aggregate cross-view context through register representations. These complementary representations form the foundation of our decoder: spatial tokens retain localized cues for \emph{where} interactions occur, while register tokens provide contextual cues for determining \emph{what} affordances are present.

\subsection{Affordance Decoder}
\label{sec:base_model}

Our affordance decoder performs prompt-free scene affordance understanding from the images used for geometric reconstruction. Unlike previous approaches ~\cite{corsetti2025fun3du,vu2026affordmatcher,su2025ovafields,schult2023mask3d,delitzas2024scenefun3d,yilmaz2026volt}, our model requires neither 3D geometry nor target queries; instead, we repurpose VGGT-$\Omega$'s representations to decode functional information, combining spatial localization with register-conditioned semantic reasoning. 

\paragraph{Affordance Discovery.}
Given $N$ images, VGGT-$\Omega$ reconstructs scene geometry and produces intermediate patch tokens and register tokens per view. We collect features from four aggregator layers\footnote{Layer 4, 11, 17, 23 following VGGT-$\Omega$'s implementation.} and project them through separate pathways. The spatial pathway fuses patch tokens using a DPT-style decoder~\cite{ranftl2021vision}, partially initialized from compatible weights of the pretrained dense head. The features are then progressively upsampled to a $512\times512$ feature map, where a lightweight RGB stem adds local image detail for fine-grained affordance localization.

The semantic pathway associates each affordance class $k \in \{1,\ldots,9\}$ with a learned query $q_k$. Two transformer-decoder layers condition these queries, one on scene-aggregated registers and the other on view-specific registers. The resulting frame-specific queries are projected against the feature map from the spatial pathway to predict nine affordance logits per pixel, along with a fixed zero background logit and softmax producing an exclusive ten-class prediction. These per-view predictions are lifted and associated across views to form 3D affordance instances.

\paragraph{Proposal Association.}
Class-wise masks localize affordances but lack instance identities. We split each class mask into 8-connected components and lift them into the reconstructed scene. Proposals are voxelized at 2\,cm and associated across views using 3D overlap, depth-consistent reprojection, and centroid agreement. Mutual-best matches initialize tracks and are progressively merged under geometric consistency and frame-uniqueness constraints to limit erroneous associations. Each track forms a 3D affordance instance from the union of its associated proposal voxels.

\paragraph{Affordance Learning.} For training, we first transfer ground truth affordance annotations from 3D scans to each RGB views using registered camera parameters and depth-aware visibility filtering. We supervise the image-space predictions using foreground-weighted multiclass cross-entropy and soft Dice loss~\cite{milletari2016vnet}. As affordance regions often occupy only a small fraction of an image, foreground weighting mitigates foreground-background imbalance, while Dice encourages accurate spatial coverage. Image-space supervision alone, however, treats repeated observations of the same physical surface independently.

Inspired by methods that aggregate predictions across views through shared 3D geometry~\cite{zhi2021semanticnerf,hu2021bpnet,mccormac2017semanticfusion}, we use reference 3D geometry during training to establish surface correspondences. Depth-aware projections are grouped into $2$\,cm surface voxels, associating pixels from different views that observe the same physical surface. Unlike methods that learn an implicit semantic field~\cite{zhi2021semanticnerf} or an additional 3D feature network~\cite{hu2021bpnet}, we use these correspondences solely to supervise image-space affordance predictions.

\begin{figure*}[t]
\centering
\qualScene
    {}
    {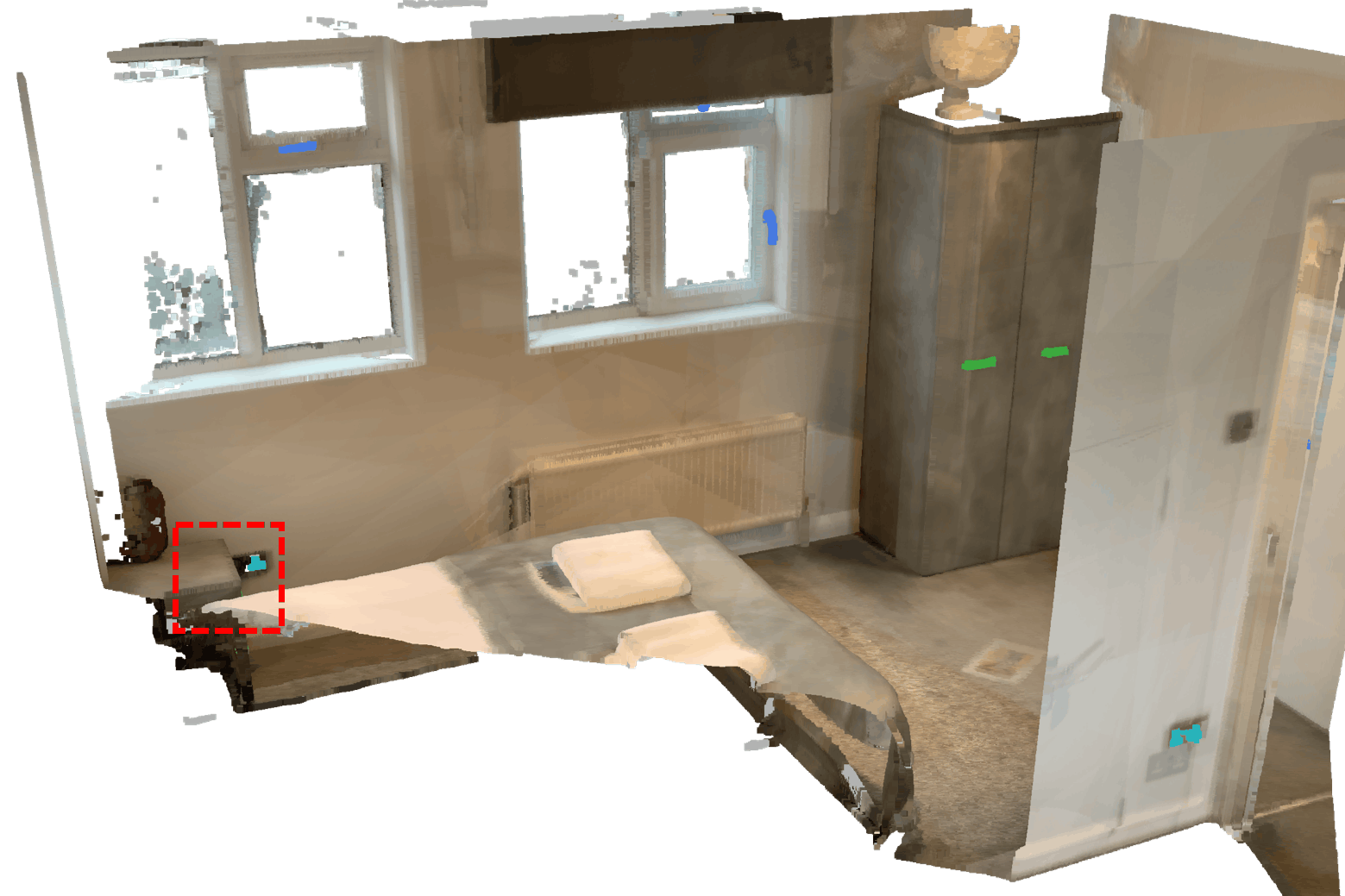}
    {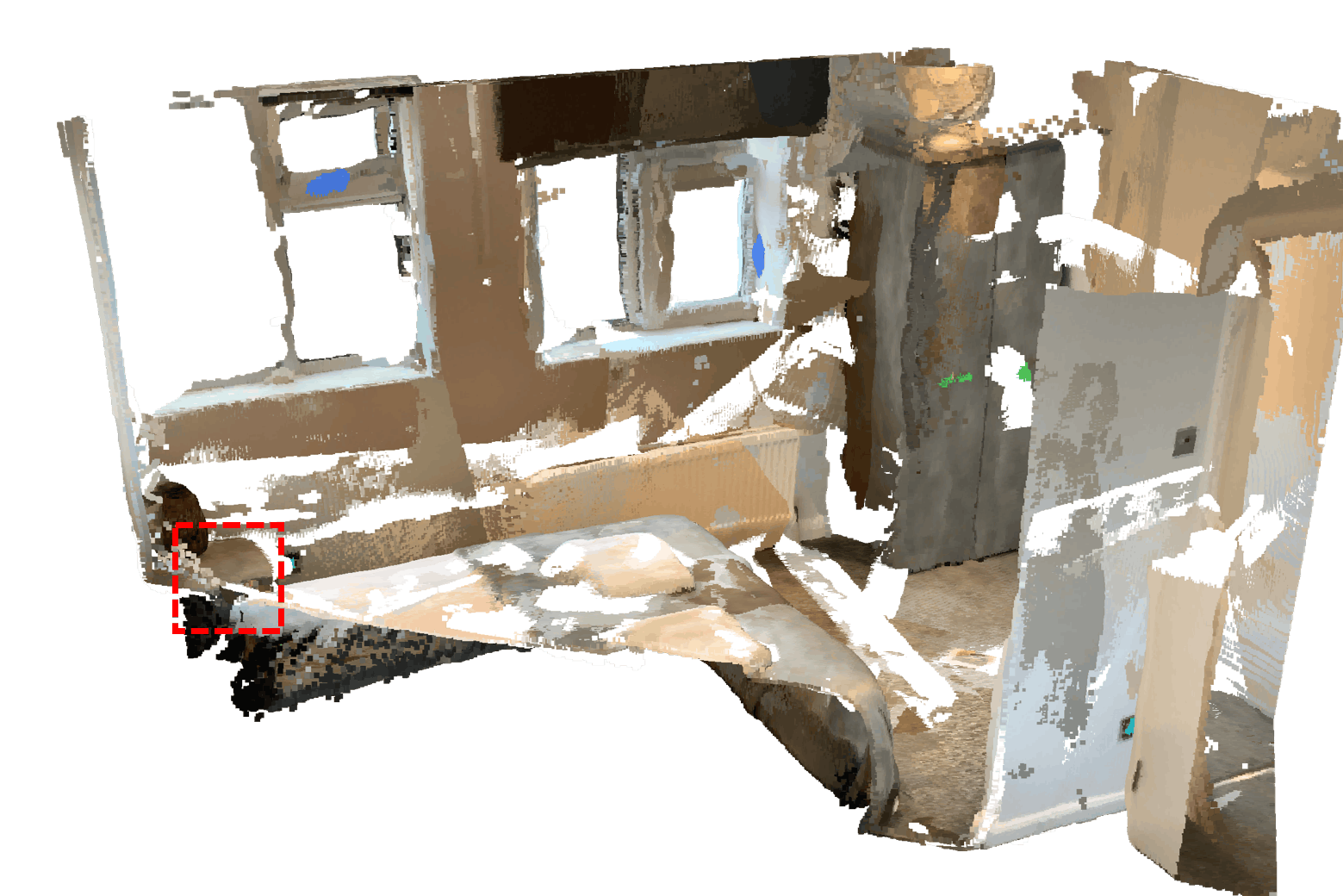}
    {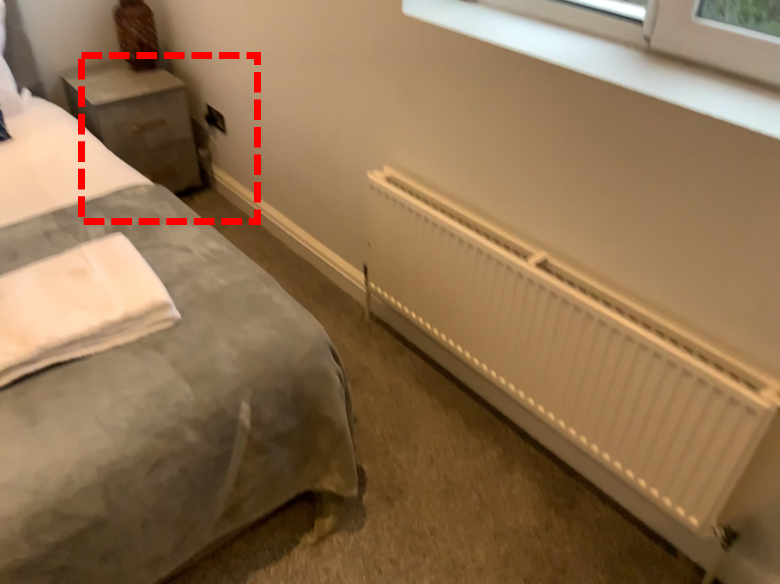}
    {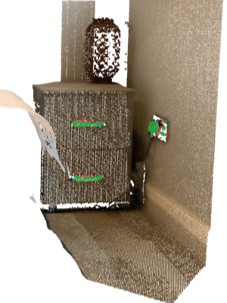}
    {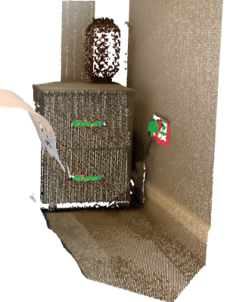}
    {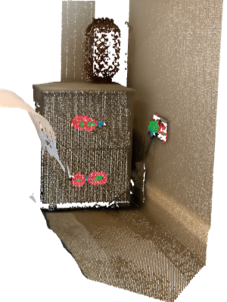}
    {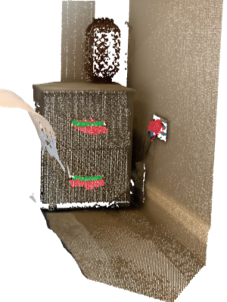}
\caption{
\textbf{Qualitative comparison.}
(top) The original point cloud and reconstructed point cloud, predicted affordances annotated. 
(bottom) Instance-level comparison of
VGGT-$\forall^\dagger$, Volt~\cite{yilmaz2026volt},
OVA-Fields~\cite{su2025ovafields},
and LERF~\cite{kerr2023lerf} on native point clouds, corresponding to the red box above.
\textcolor{green!60!black}{Green},
\textcolor{red}{red}, and
\textcolor{blue}{blue} denote true positives,
false positives, and false negatives. As shown, VGGT-$\forall^\dagger$ provides a wide affordance coverage while minimizing false positives. We provide more visual examples in the \textbf{Supplementary Material}.
}
\vspace{-16pt}

\label{fig:qualitative_1}
\end{figure*}

Let $\mathcal{O}_v$ denote the observations associated with surface voxel $v$, and let $\bar{\mathbf p}_v$ represent their mean predicted semantic distribution. Collecting these predictions as $\bar{\mathbf P}$ with ground-truth labels $\mathbf Y$, we define the surface-level objective:

\begin{equation}
\mathcal{L}_{\mathrm{surf}}
=
\mathcal{L}_{\mathrm{CE}}(\bar{\mathbf P},\mathbf Y)
+
\mathcal{L}_{\mathrm{Dice}}(\bar{\mathbf P},\mathbf Y)
\label{eq:loss_surface}
\end{equation}
This objective supervises affordance semantics on shared physical surfaces rather than isolated image observations.

A correct surface-level average, however, does not guarantee consistency among its individual observations. Unlike SegVGGT's frame-level query-attention alignment~\cite{qu2026segvggt}, we enforce semantic agreement across observations of the same physical surface. We define $\mathbf c_i$ for each prediction $\mathbf p_i$ as the detached mean prediction of observations from other views corresponding to the same surface. We then minimize
\begin{equation}
\mathcal{L}_{\mathrm{cv}}
=
\frac{1}{|\mathcal{C}|}
\sum_{i\in\mathcal{C}}
\operatorname{JS}
\left(
\mathbf p_i \,\|\, \mathbf c_i
\right)
\label{eq:loss_cv}
\end{equation}
where $\mathcal{C}$ contains observations of surfaces visible in at least two distinct views. The leave-one-view-out consensus excludes the current observation from its target, while detachment prevents gradients from propagating through the target, encouraging consistent predictions across viewpoints.

Finally, we combine the image-space, surface-level, and cross-view objectives as
\begin{equation}
\mathcal{L}
=
\underbrace{
\mathcal{L}_{\mathrm{CE}}
+
\mathcal{L}_{\mathrm{Dice}}
}_{\mathcal{L}_{2D}}
+
\lambda_\mathrm{surf}\,\mathcal{L}_{\mathrm{surf}}
+
\lambda_\mathrm{cv}\,\mathcal{L}_{\mathrm{cv}}
\label{eq:total_loss}
\end{equation}
\
$\mathcal{L}_{\mathrm{surf}}$ anchors each shared physical region to its affordance label, while $\mathcal{L}_{\mathrm{cv}}$ regularizes its individual observations toward a consistent prediction. Reference geometry provides supervision only during training; at inference, VGGT-$\forall$ reconstructs both geometry and affordances from uncalibrated RGB images alone.

\subsection{\textbf{VGGT-$\forall^\dagger$}}
\label{sec:scan_assisted}

While VGGT-$\forall$ discovers 3D affordances from images alone, established benchmarks~\cite{delitzas2024scenefun3d} evaluate predictions on high-fidelity native point clouds. As also observed by SegVGGT~\cite{qu2026segvggt}, transferring image-derived predictions onto reference geometry introduces projection errors and misalignments, potentially degrading 3D segmentation accuracy. To address this, we introduce \textbf{VGGT-$\forall^\dagger$}, a scan-assisted variant that uses VGGT-$\forall$'s results to guide localized 3D segmentation on the 3D scan. Figure~\ref{fig:main_figure} illustrates both VGGT-$\forall$ and VGGT-$\forall^\dagger$.

We first align the predicted affordances with the point cloud through geometric correspondences, without ground-truth masks. Each affordance defines a local point-cloud crop with 0.08\,m horizontal padding spanning the full scan height. To adapt existing 3D segmentation architectures~\cite{zhou2025pointsam,choy2026spaceformer,yilmaz2026volt} towards the cropped local regions, we use rank-64 LoRA~\cite{hu2022lora} adapters in the transformer layers and a trainable semantic head. Training uses native pointwise annotations with balanced sampling for affordance classes and prediction, along with gravity-axis rotational augmentation. We combine cross-entropy for pointwise class discrimination with Lovász loss~\cite{berman2018lovasz} to optimize for small affordance regions. By predicting all affordances and background within each crop, the adapted model can reject false detections and correct initial class assignments.

At inference, we follow Volt~\cite{yilmaz2026volt} by applying test-time augmentation and voxel voting to each candidate crop. The resulting predictions are mapped to native scan vertices, with overlapping instances consolidated and reranked using geometric consistency and VGGT-$\forall$ support. This enables competitive native-point segmentation through localized refinement rather than full-scene segmentation inference.

\section{Experiments}
\label{sec:experiments}


We evaluate VGGT-$\forall$ and VGGT-$\forall^\dagger$ under two complementary settings. The established SceneFun3D benchmark~\cite{scenefun3d_task1} evaluates affordance on laser scanned point clouds, and cannot assess RGB-only affordance discovery. We therefore establish a new RGB-only evaluation protocol on image-reconstructed geometry, transferring annotations through RGB-view correspondences and evaluating all adapted methods under a common reconstructed-surface protocol. VGGT-$\forall^\dagger$ is evaluated under the official native point-cloud protocol, including the hidden test set~\cite{scenefun3d_leaderboard}, to assess our scan-assisted alignment of image-discovered affordances to an established benchmark. Sec.~\ref{sec:exp_setup} describes the setup, Secs.~\ref{sec:exp_native} and \ref{sec:exp_base} report the two evaluations, and Sec.~\ref{sec:exp_ablation} presents ablation studies. Further details and experiments are provided in the \textbf{Supplementary Material}.

\subsection{Experimental Setup}
\label{sec:exp_setup}

\paragraph{Data Processing.}
SceneFun3D~\cite{delitzas2024scenefun3d} provides 200 training, 30 validation, and 85 hidden-test scenes with affordance instances annotated on native 3D scans and corresponding RGB-D observations. For VGGT-$\forall$, we project 3D annotations onto RGB frames using depth-aware visibility filtering. We then construct 12-view samples selected to cover diverse viewpoints, requiring each included class to appear in at least four views. The resulting RGB-only manifests contain 2,200 training samples from 188 scenes and 311 validation windows from 26 scenes, with some scenes excluded due to projection error. Same-class instances are merged into semantic masks for training, while instance identities are retained for evaluation. Images and masks are processed at $512\times512$, with registered 2\,cm surface voxel establishing cross-view correspondences. For VGGT-$\forall^\dagger$, we construct 10,417 VGGT-$\forall$-guided native-point crops with $0.08$\,m XY padding, spanning the full scan height.

\paragraph{Model Implementation.}
For VGGT-$\forall$, we freeze the VGGT-$\Omega$ trained at $512\times512$ resolution and train a 39M-parameter affordance decoder, initializing compatible spatial layers from its pretrained dense head. The decoder is optimized using Eq.~\ref{eq:total_loss}, with $\lambda_\text{surf} = 0.1$ and $\lambda_{cv} = 0.5$, respectively. Reference geometry and camera parameters support training supervision and evaluation; inference for VGGT-$\forall$ requires RGB images alone. For VGGT-$\forall^\dagger$,  we evaluate pretrained 3D segmentation models such as Point-SAM~\cite{zhou2025pointsam}, SpaceFormer~\cite{choy2026spaceformer}, and Volt~\cite{yilmaz2026volt}, selecting Volt as our default refiner. We use a rank-64 LoRA adapter and a trainable semantic head, while keeping the base network frozen. At inference, a native 3D scan with its registered RGB-D observations and camera poses are provided as inputs. Per-view masks predicted by VGGT-$\forall$ are projected onto the 3D scan to identify native-point crops for localized segmentation. All training experiments are conducted on a single workstation with four NVIDIA RTX A5000 GPUs.

\begin{table}[t]
\centering
\caption{
Native point-cloud evaluation on SceneFun3D and Articulate3D validation sets.
Best results are bold, second-best underlined.
Blue superscripts indicate gains over the best prior method.
}
\vspace{-6pt}
\label{tab:scenefun3d_val}

\begingroup
\fontsize{8}{9.5}\selectfont
\setlength{\tabcolsep}{2.5pt}

\newcommand{\gain}[1]{%
  \textsuperscript{\textcolor{blue!70!black}{\scriptsize +#1}}%
}

\resizebox{\columnwidth}{!}{%
\begin{tabular}{@{}lcccccc@{}}
\toprule

\multirow{2}{*}{Method}
& \multicolumn{3}{c}{Precision}
& \multicolumn{3}{c}{Recall} \\
\cmidrule(lr){2-4}
\cmidrule(l){5-7}

& $\mathrm{mAP}$ $\uparrow$
& $\mathrm{mAP}_{50}$ $\uparrow$
& $\mathrm{mAP}_{25}$ $\uparrow$
& $\mathrm{mAR}$ $\uparrow$
& $\mathrm{mAR}_{50}$ $\uparrow$
& $\mathrm{mAR}_{25}$ $\uparrow$ \\

\midrule
\multicolumn{7}{@{}l}{%
\textit{SceneFun3D Validation (9 classes)}
} \\
\midrule

LERF~\cite{kerr2023lerf} {\scriptsize(ICCV'23)}
& 2.0 & 4.6 & 26.7 & 9.2 & 20.8 & 53.7 \\

Mask3D-F~\cite{delitzas2024scenefun3d,  schult2023mask3d} {\scriptsize(CVPR'24)}
& 7.9 & 13.6 & 18.8 & 15.3 & 25.6 & 34.2 \\

OVA-Fields~\cite{su2025ovafields} {\scriptsize(ICCV'25)}
& 1.3 & 5.0 & 36.5 & 7.0 & 19.6
& 57.9 \\

Volt~\cite{yilmaz2026volt} {\scriptsize(ECCV'26)}
& \underline{39.6}
& \underline{53.6}
& \underline{57.7}
& \underline{44.5}
& \underline{58.1}
& \underline{62.3} \\

\midrule

VGGT-$\forall^\dagger$ (Ours)
& \textbf{40.4}\gain{0.8}
& \textbf{54.4}\gain{0.8}
& \textbf{60.8}\gain{3.1}
& \textbf{46.8}\gain{2.3}
& \textbf{61.1}\gain{3.0}
& \textbf{67.2}\gain{4.9} \\

\midrule
\multicolumn{7}{@{}l}{%
\textit{Articulate3D Validation (3 classes)}
} \\
\midrule

Volt~\cite{yilmaz2026volt} {\scriptsize(ECCV'26)}
& \underline{17.8}
& \underline{26.5}
& \underline{33.1}
& \underline{21.5}
& \underline{29.8}
& \underline{35.4} \\

VGGT-$\forall^\dagger$ (Ours)
& \textbf{18.8}\gain{1.0}
& \textbf{26.7}\gain{0.2}
& \textbf{33.6}\gain{0.5}
& \textbf{22.9}\gain{1.4}
& \textbf{30.5}\gain{0.7}
& \textbf{37.7}\gain{2.3} \\

\bottomrule
\end{tabular}%
}

\endgroup
\vspace{-6pt}
\end{table}

\begin{table}[t]
\centering
\caption{
SceneFun3D functionality segmentation on the hidden test set.
All methods are evaluated on native 3D point clouds.
}
\vspace{-6pt}
\label{tab:scenefun3d_test}
\resizebox{\columnwidth}{!}{
\fontsize{7}{9}\selectfont
\begin{tabular}{lccc}
\toprule
Method
& $\mathrm{mAP}$ $\uparrow$
& $\mathrm{mAP}_{50}$ $\uparrow$
& $\mathrm{mAP}_{25}$ $\uparrow$ \\
\midrule
LERF~\cite{kerr2023lerf}
& 4.8 & 12.3 & 18.1 \\
OpenTrack3D~\cite{zhou2026opentrack3d}
& 6.5 & 14.0 & 27.8 \\
Mask3D-F~\cite{delitzas2024scenefun3d, schult2023mask3d}
& 7.9 & 18.3 & 26.6 \\
Volt~\cite{yilmaz2026volt}
& 22.7 & \textbf{34.5} & 42.6 \\
\midrule
\textbf{VGGT-$\forall^{\dagger}$}
& \textbf{23.0} & \textbf{34.5} & \textbf{44.2} \\
\bottomrule
\end{tabular}
}
\vspace{-6pt}
\end{table}

\subsection{Scanned Point-Cloud Evaluation}
\label{sec:exp_native}
\begin{figure*}[t]
\centering

\begin{tabular*}{\textwidth}{
    @{\extracolsep{\fill}}ccccc@{}
}

\multicolumn{5}{@{}l}{
    \textbf{Scene 1: Bedroom (Affordance Classes : Hook Turn, Hook Pull)}
} \\[2pt]

\includegraphics[width=0.19\textwidth,height=2.5cm]
    {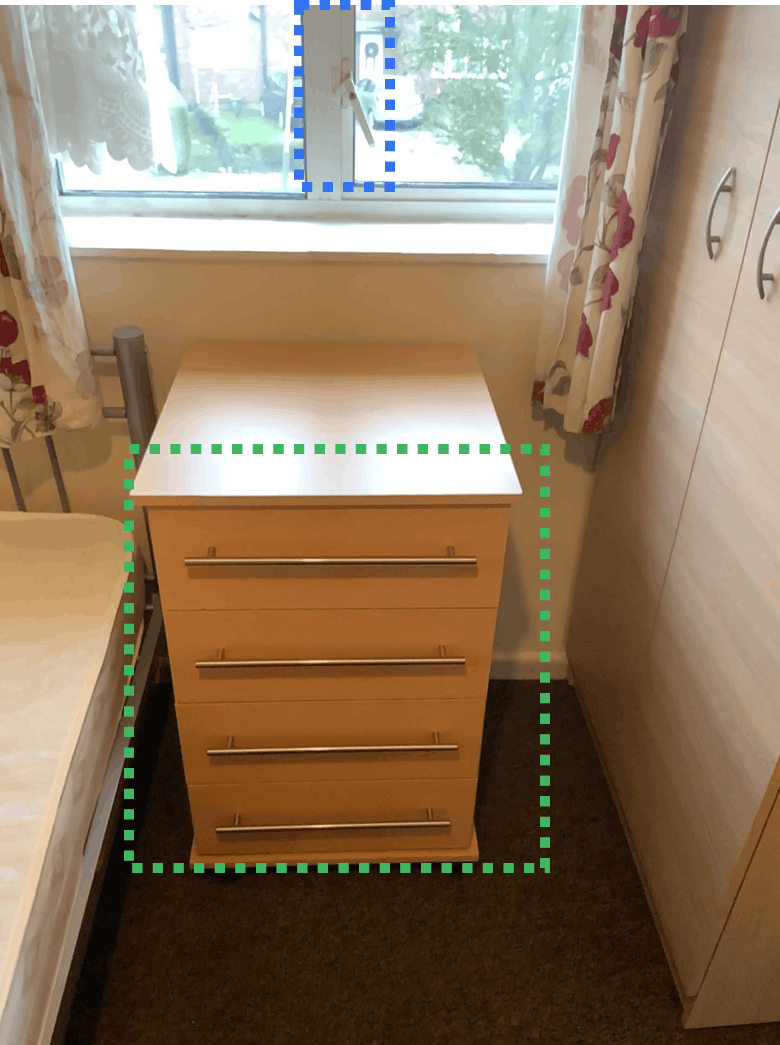} &
\includegraphics[width=0.19\textwidth,height=2.5cm]
    {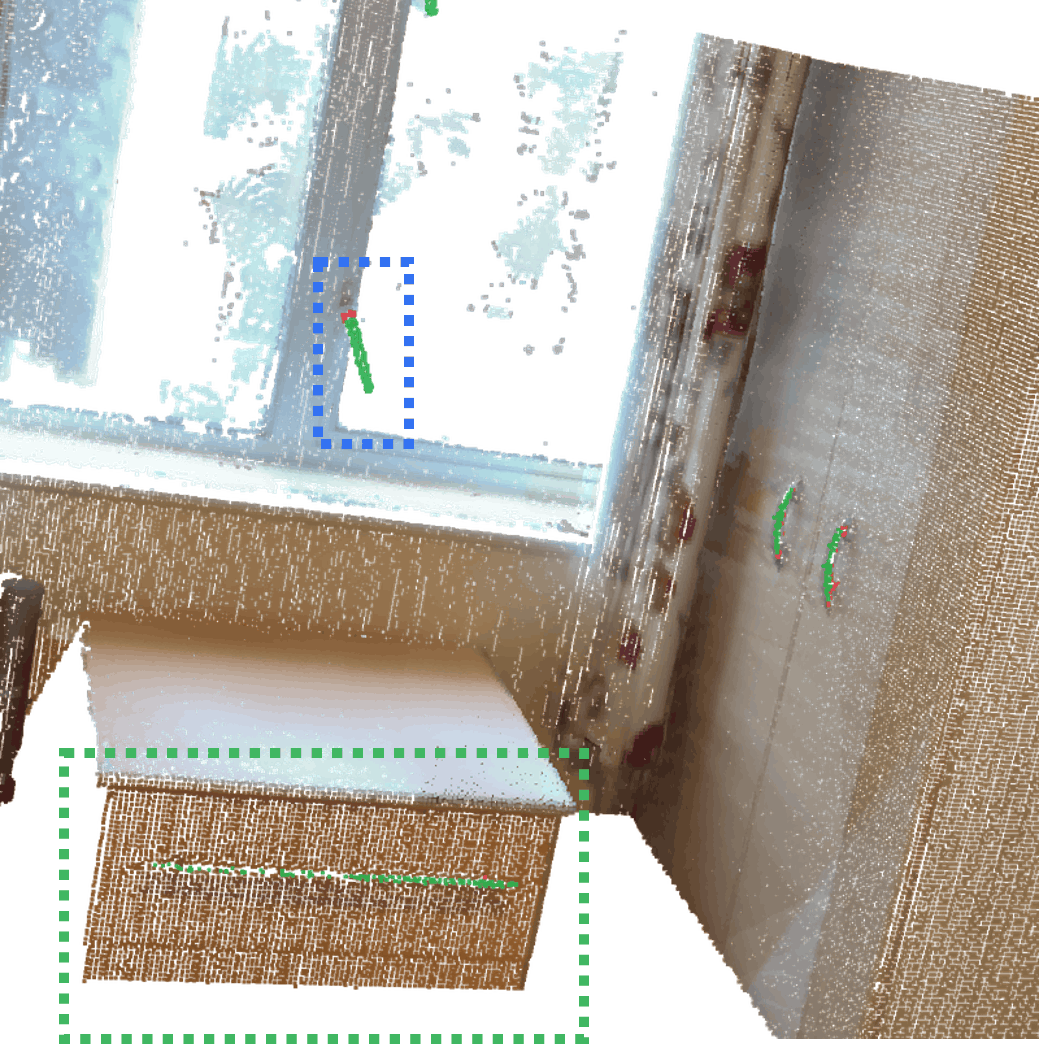} &
\includegraphics[width=0.19\textwidth,height=2.5cm]
    {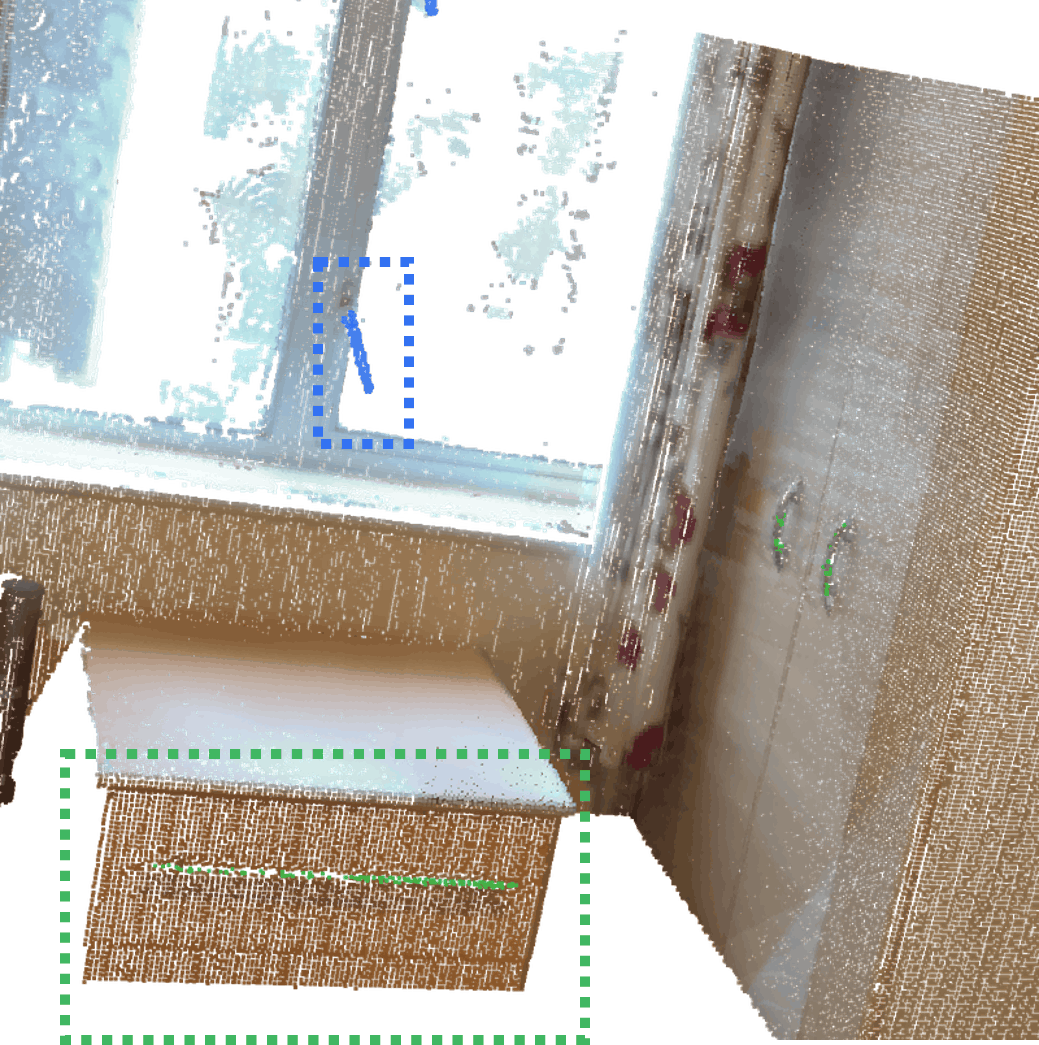} &
\includegraphics[width=0.19\textwidth,height=2.5cm]
    {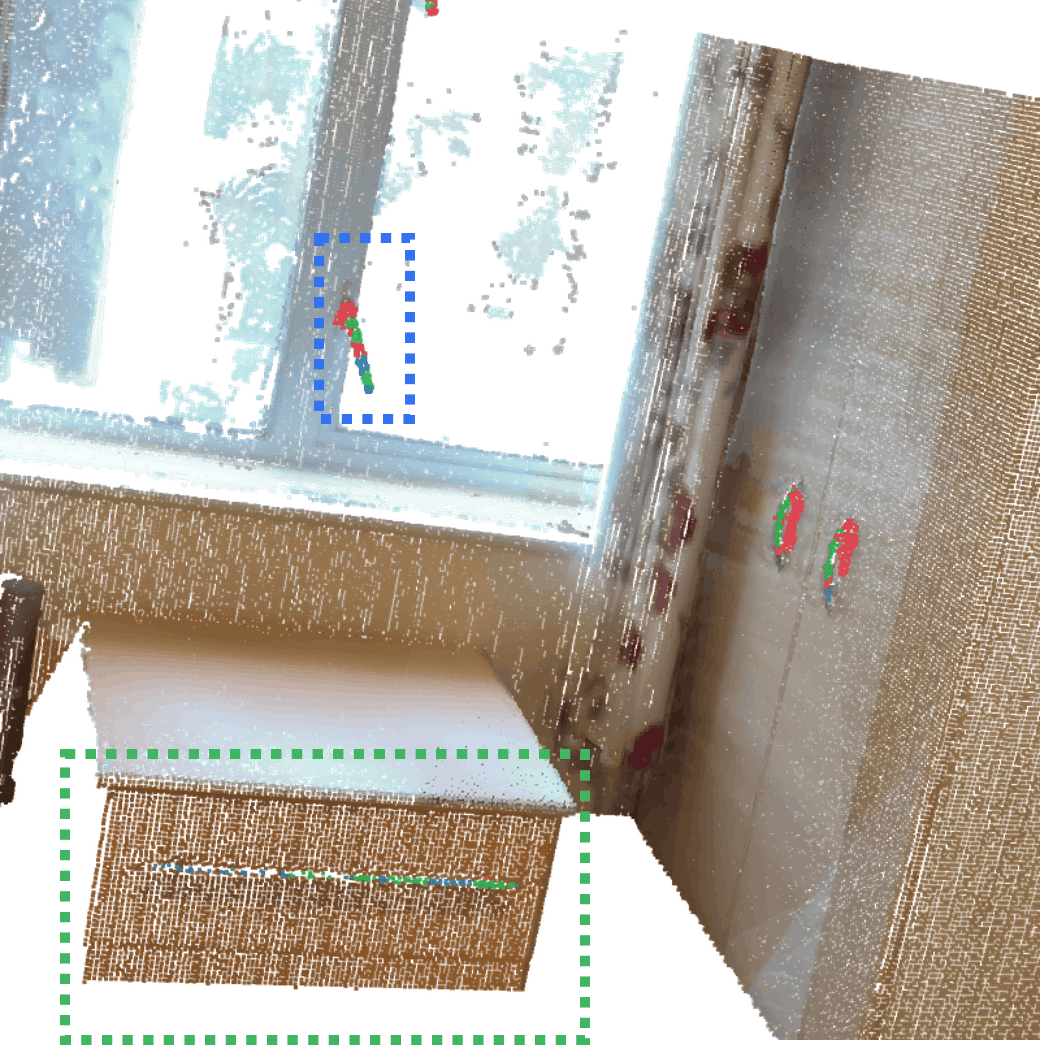} &
\includegraphics[width=0.19\textwidth,height=2.5cm]
    {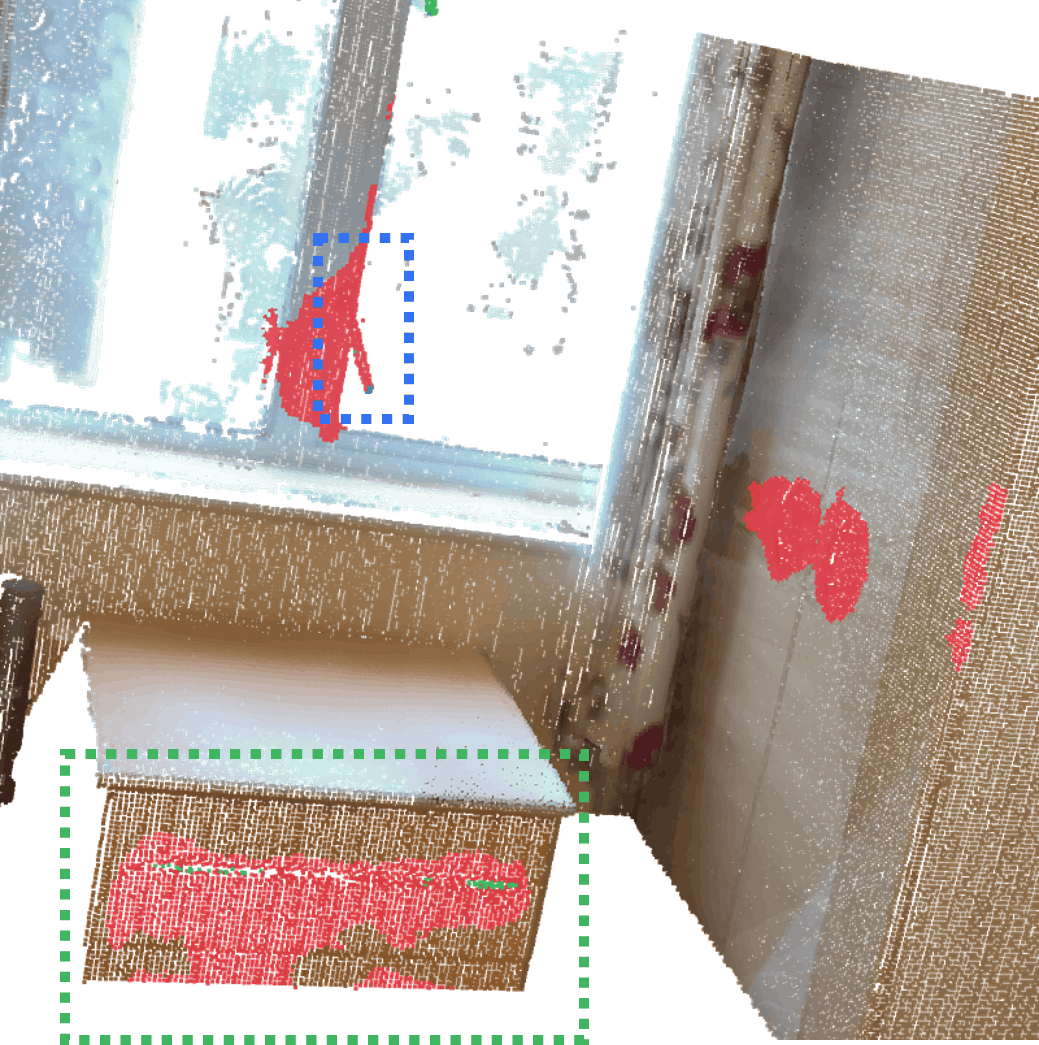}
\\[4pt]

\multicolumn{5}{@{}l}{
    \textbf{Scene 2: Bathroom (Affordance Classes : Rotate, Foot Push)}
} \\[2pt]

\includegraphics[width=0.19\textwidth,height=2.5cm]
    {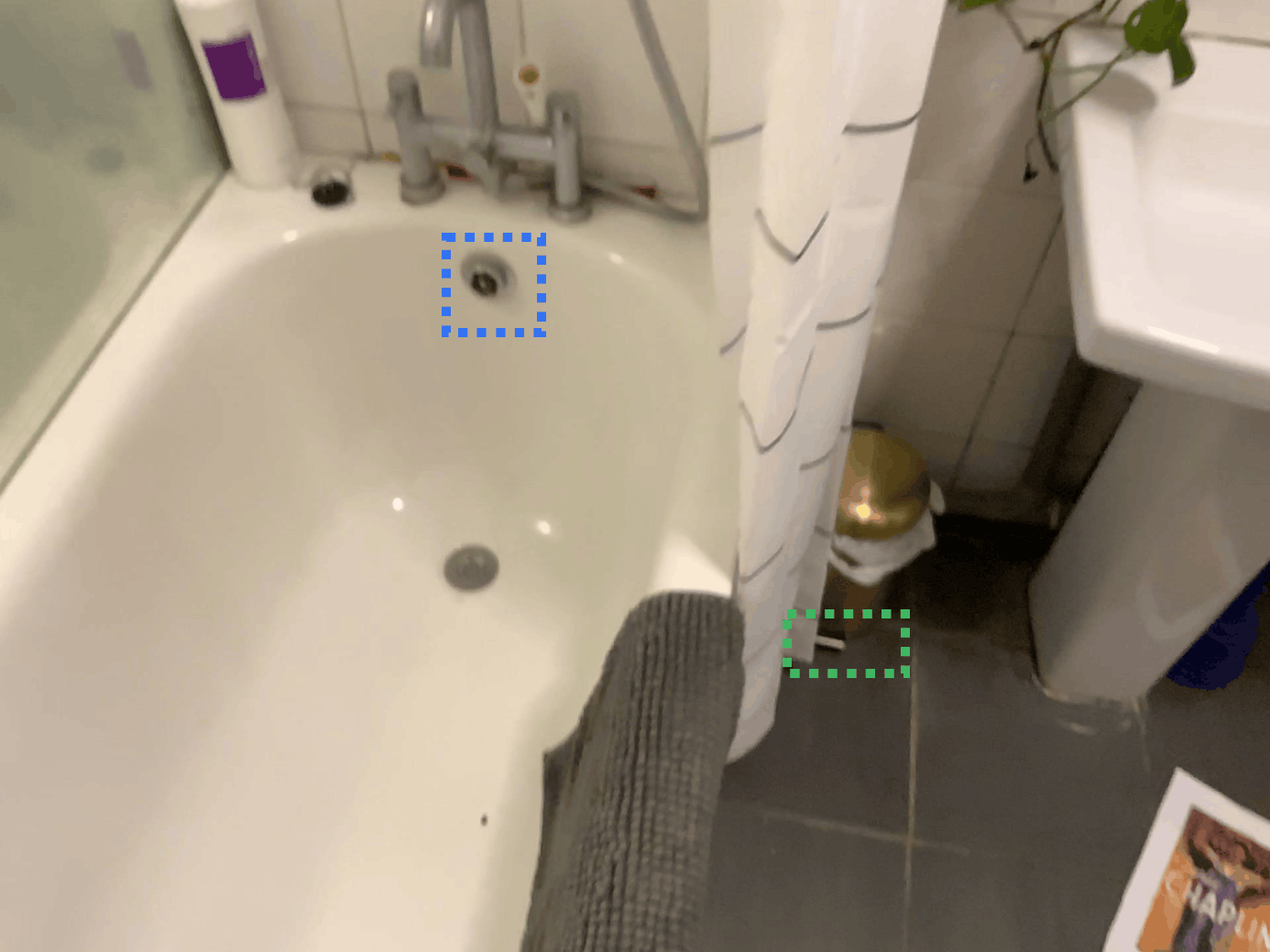} &
\includegraphics[width=0.19\textwidth,height=2.5cm]
    {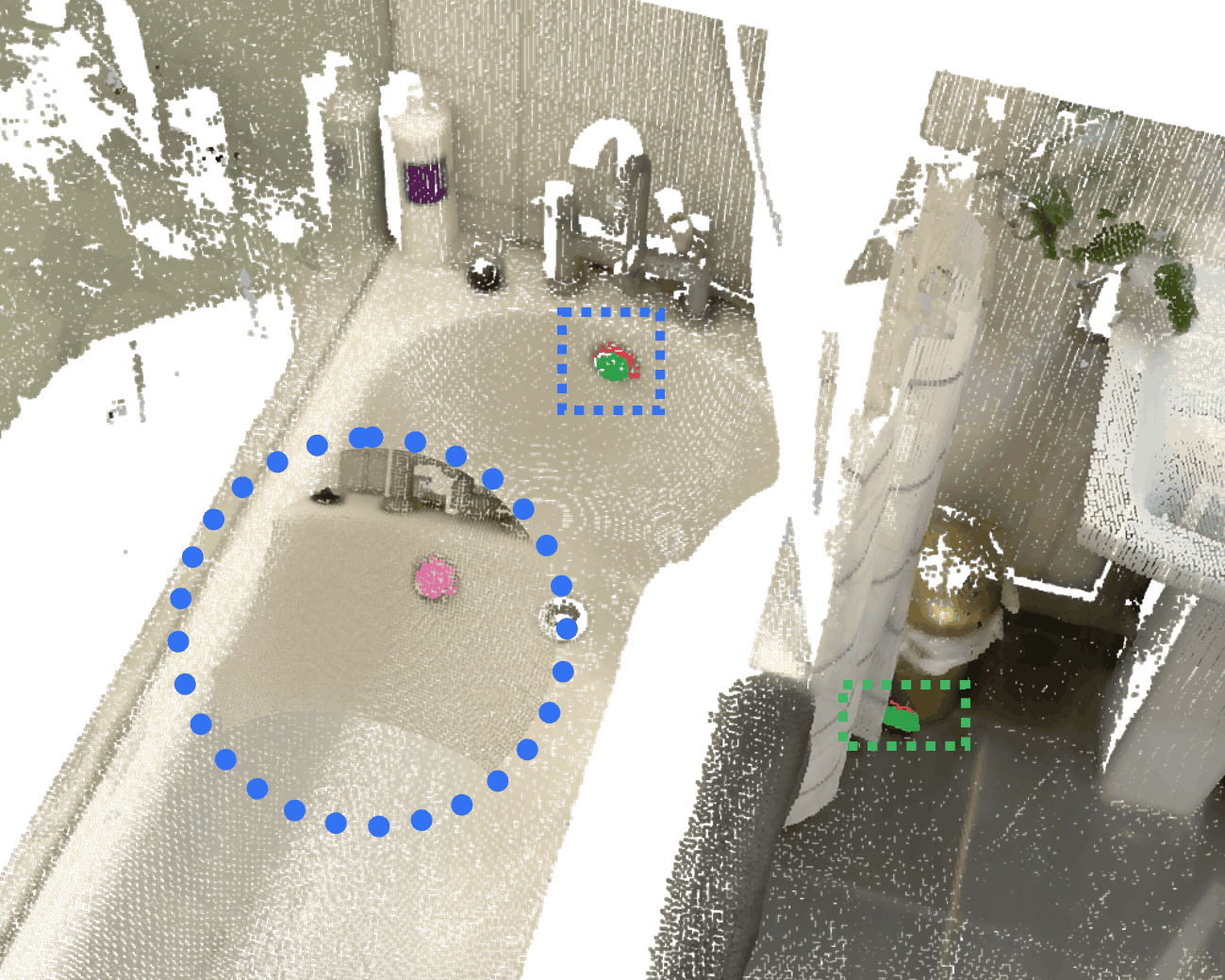} &
\includegraphics[width=0.19\textwidth,height=2.5cm]
    {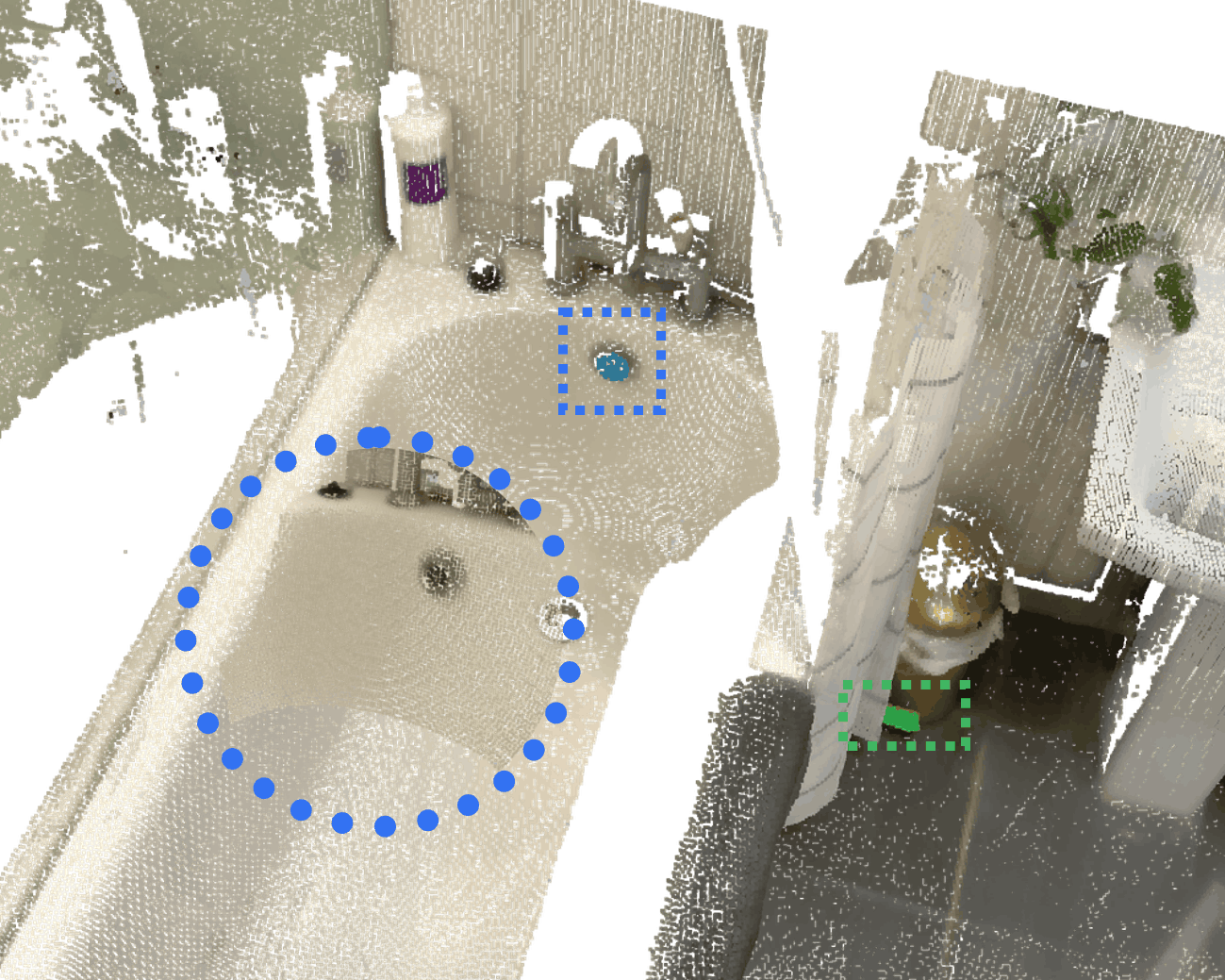} &
\includegraphics[width=0.19\textwidth,height=2.5cm]
    {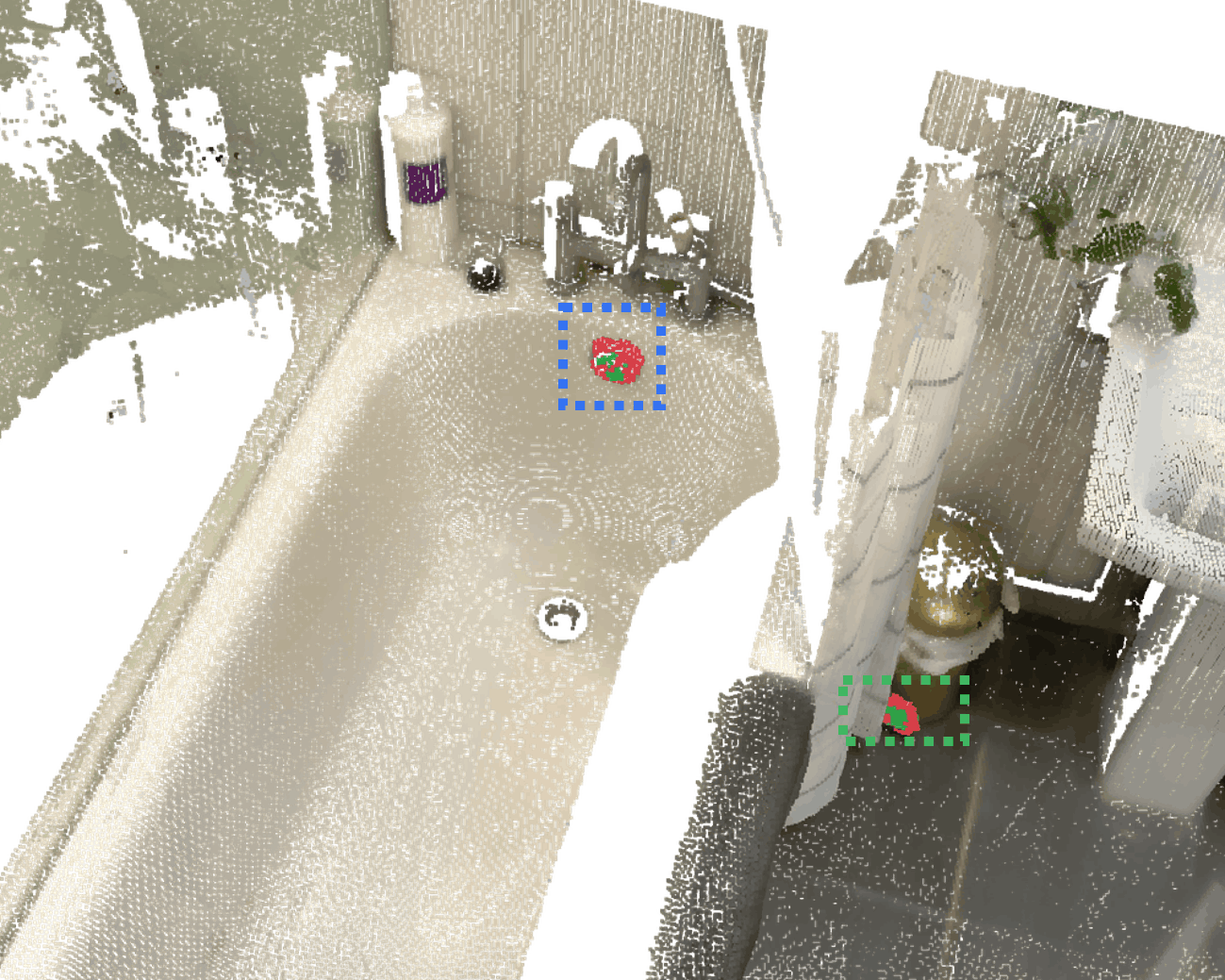} &
\includegraphics[width=0.19\textwidth,height=2.5cm]
    {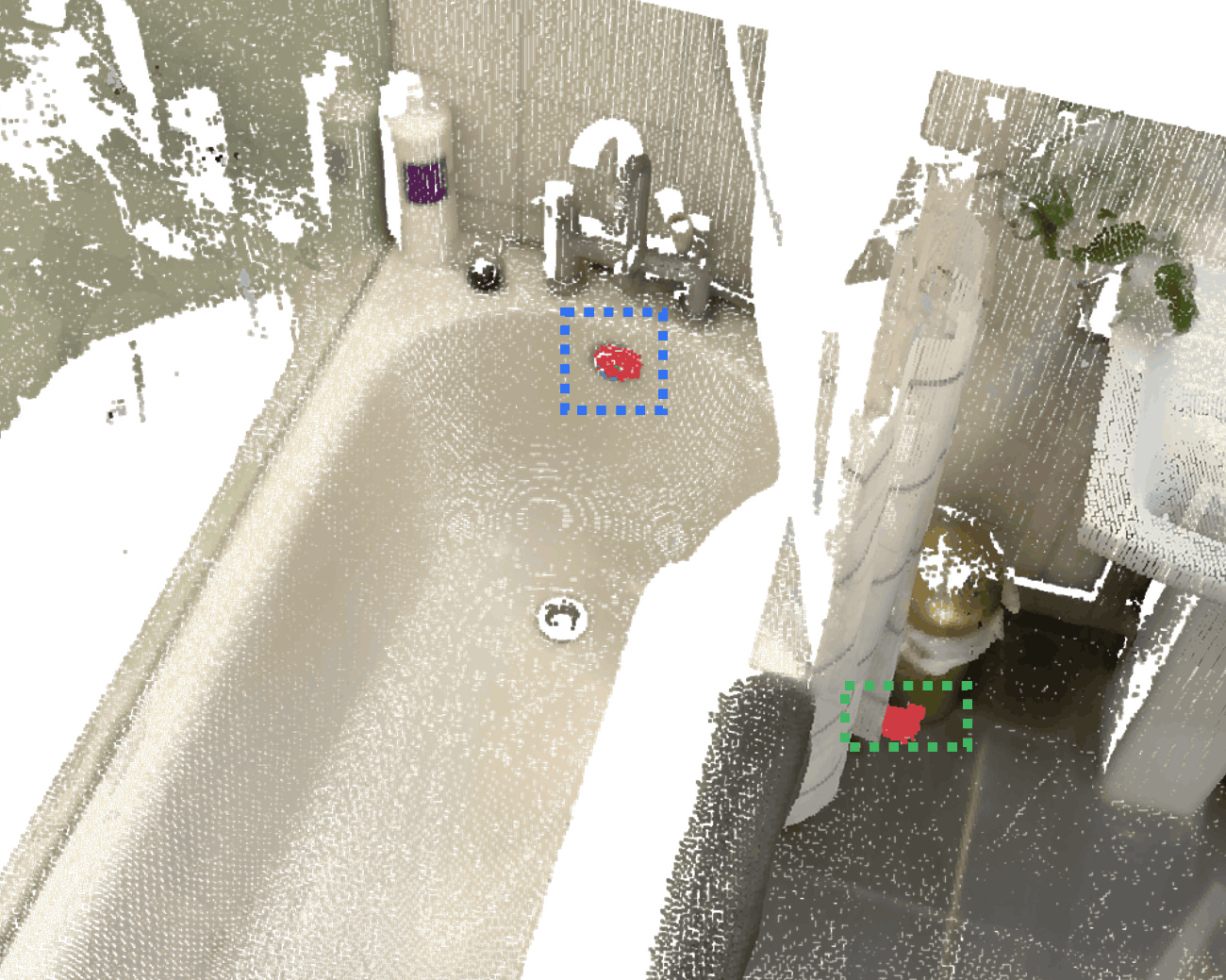}
\\[4pt]

\multicolumn{5}{@{}l}{
    \textbf{Scene 3: Bedroom (Affordance Classes : Plug In, Unplug, Pinch Pull)}
} \\[2pt]

\includegraphics[width=0.19\textwidth,height=2.5cm]
    {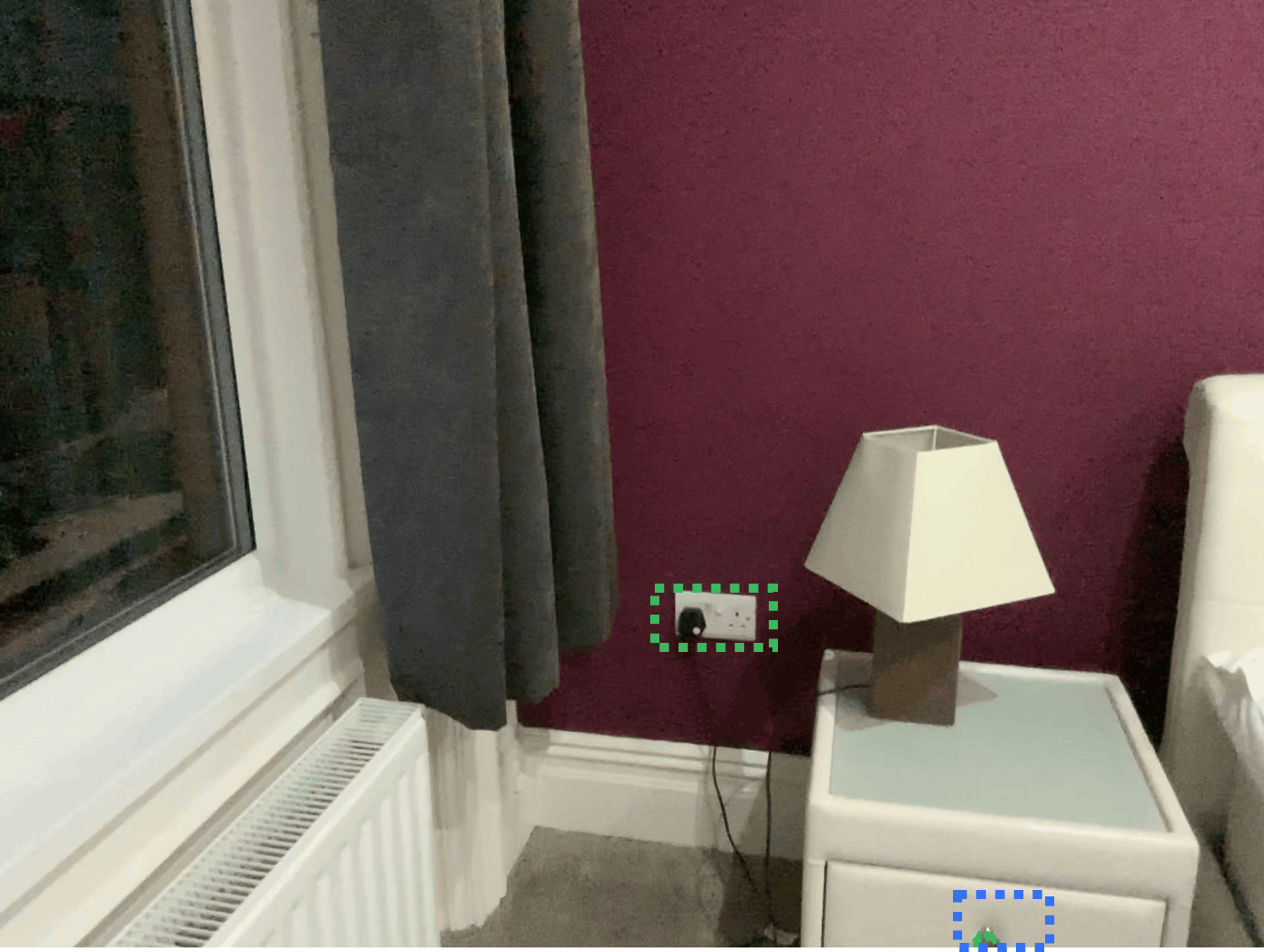} &
\includegraphics[width=0.19\textwidth,height=2.5cm]
    {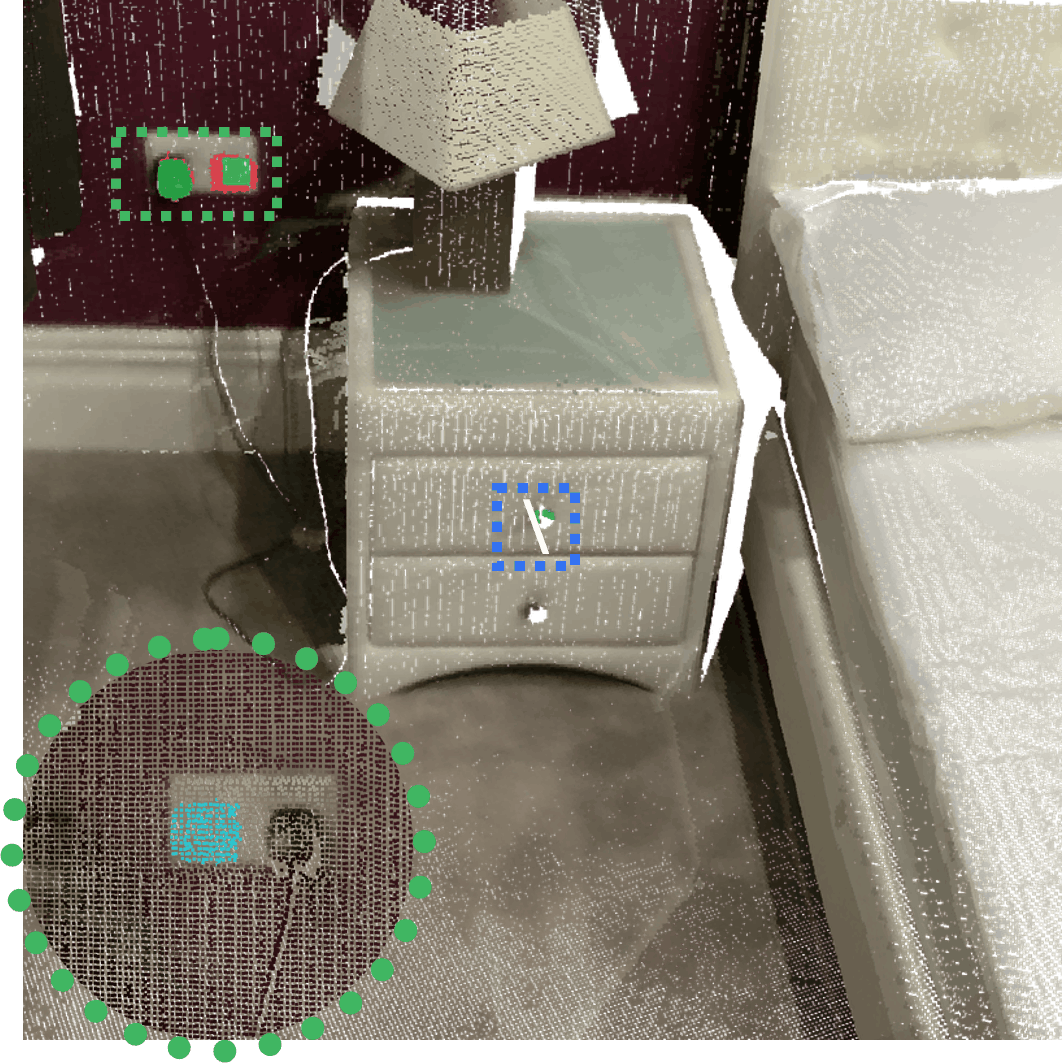} &
\includegraphics[width=0.19\textwidth,height=2.5cm]
    {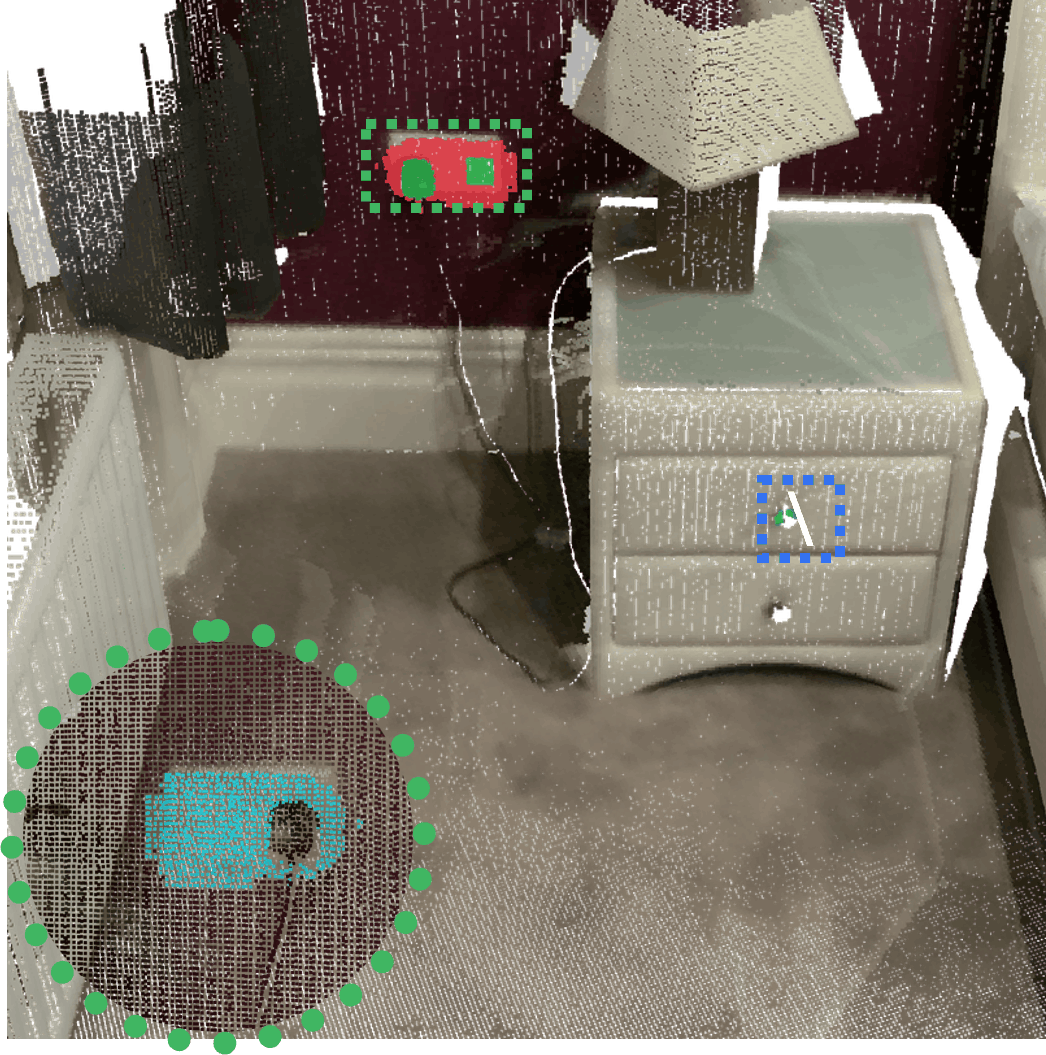} &
\includegraphics[width=0.19\textwidth,height=2.5cm]
    {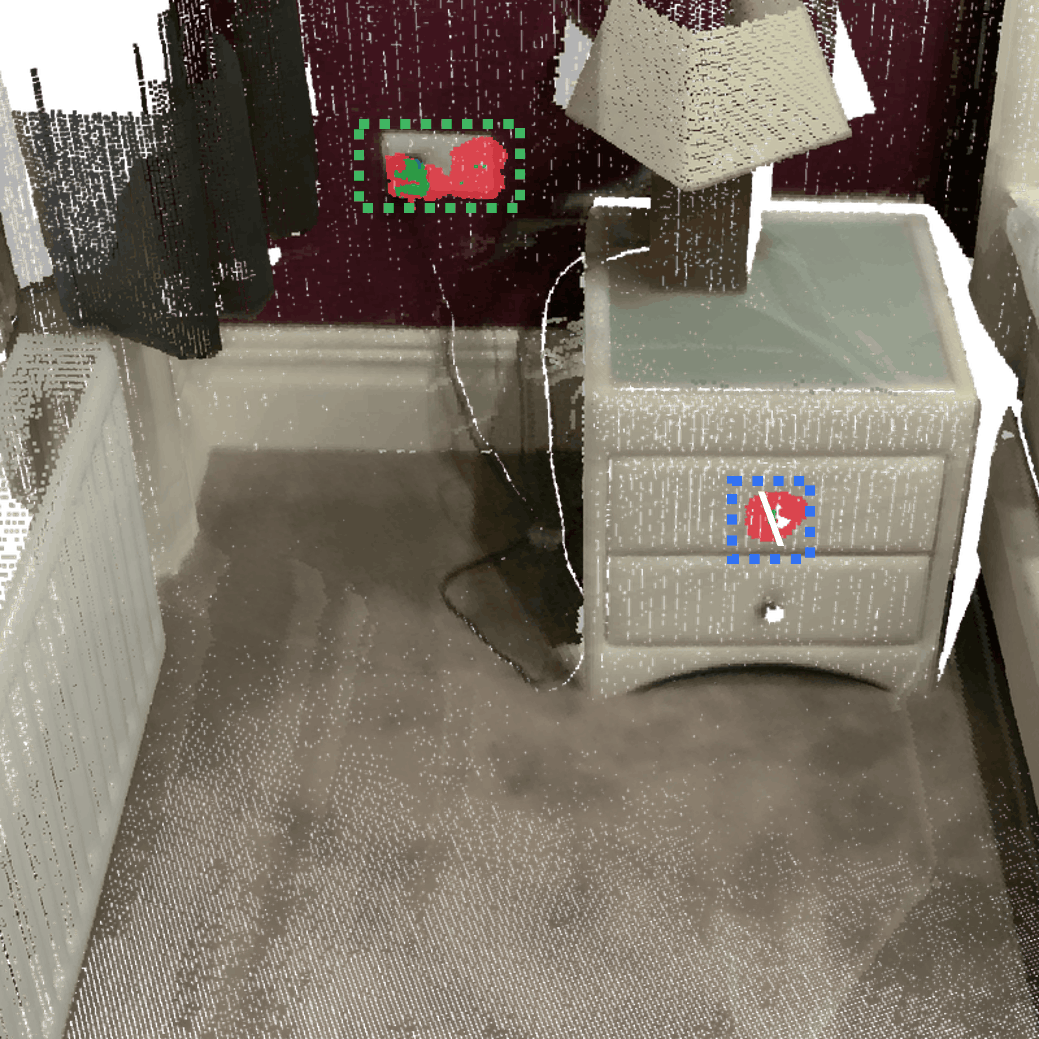} &
\includegraphics[width=0.19\textwidth,height=2.5cm]
    {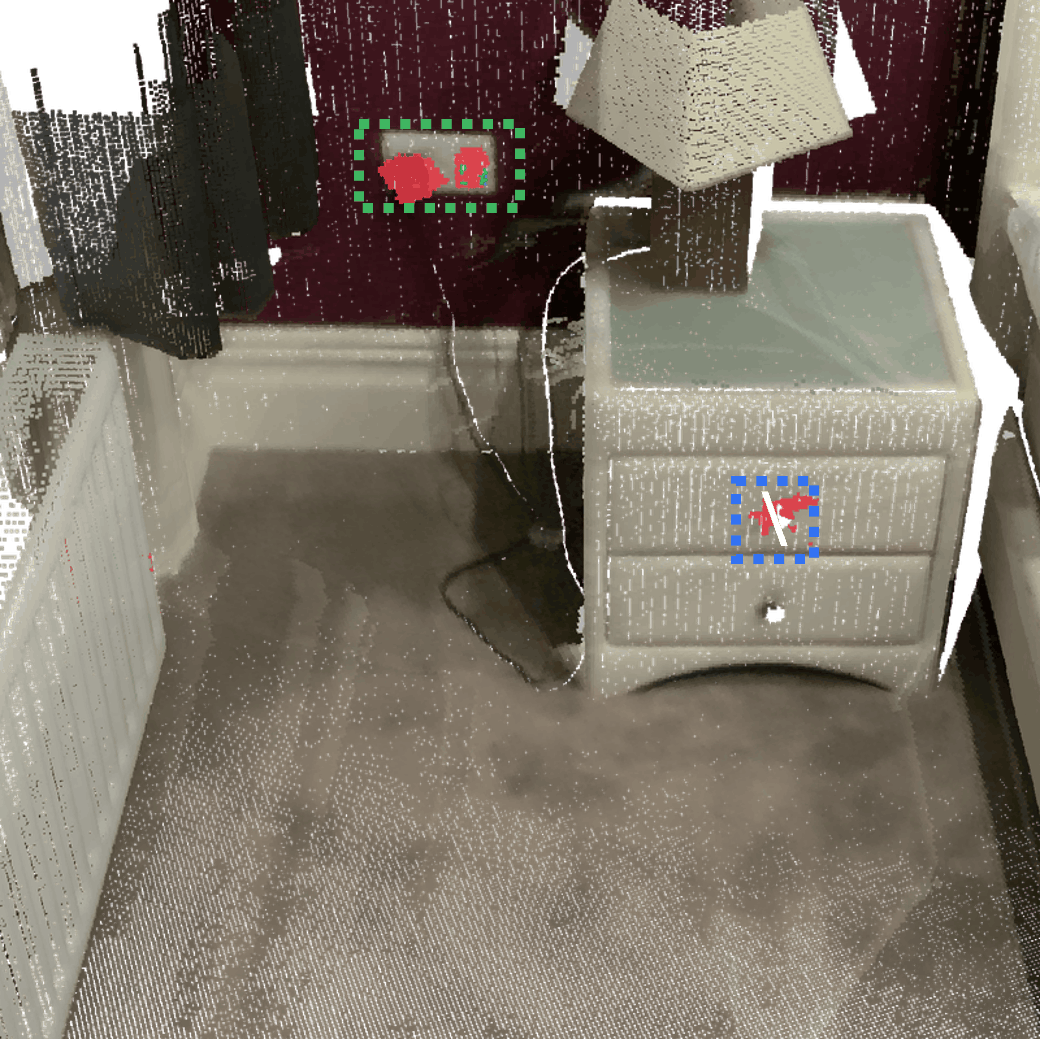}
\\[2pt]

\multicolumn{5}{@{}l}{
    \textbf{Scene 4: Living Room (Affordance Classes : Key Press, Hook Pull, Rotate)}
} \\[2pt]

\includegraphics[width=0.19\textwidth,height=2.5cm]
    {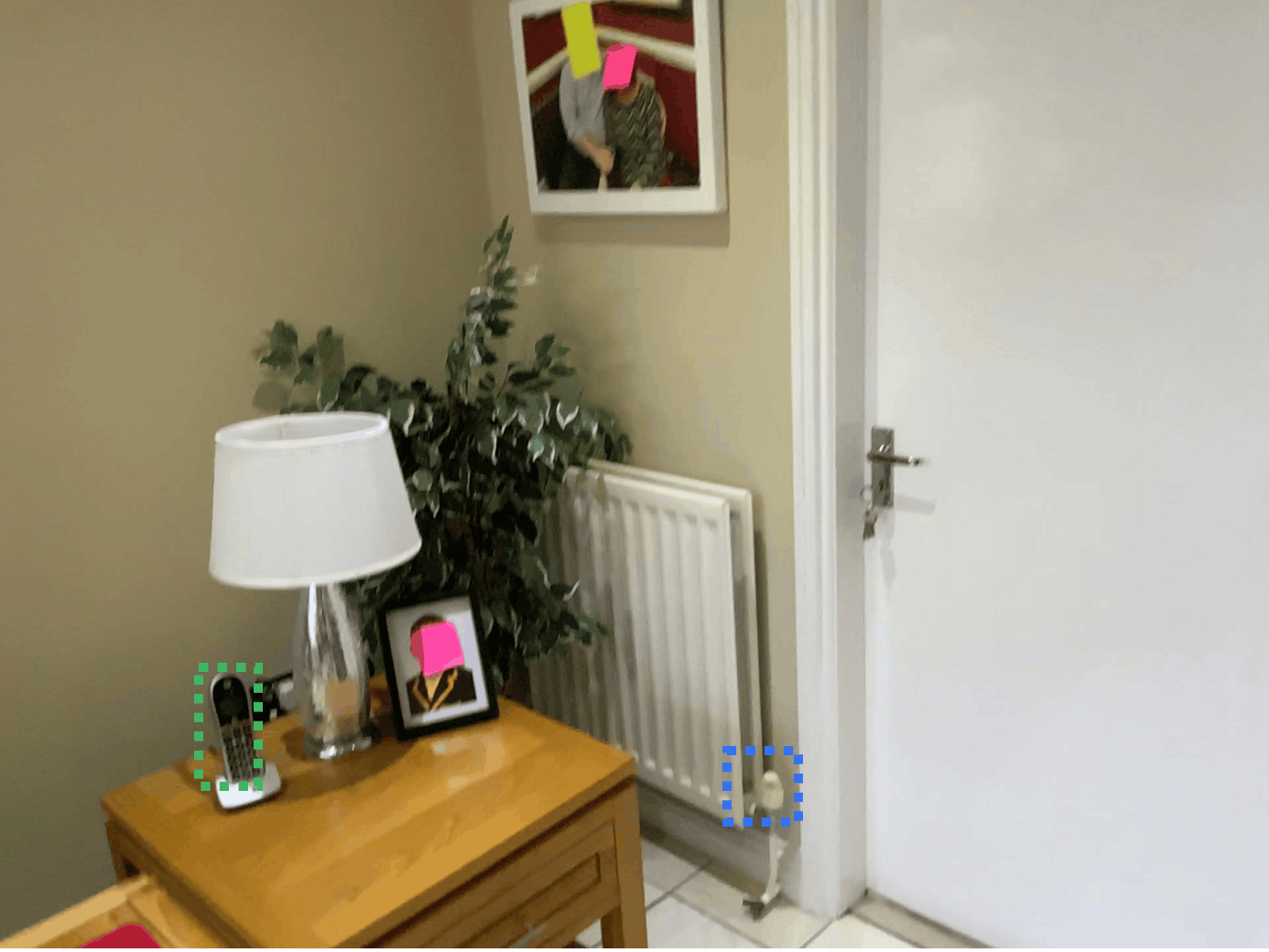} &
\includegraphics[width=0.19\textwidth,height=2.5cm]
    {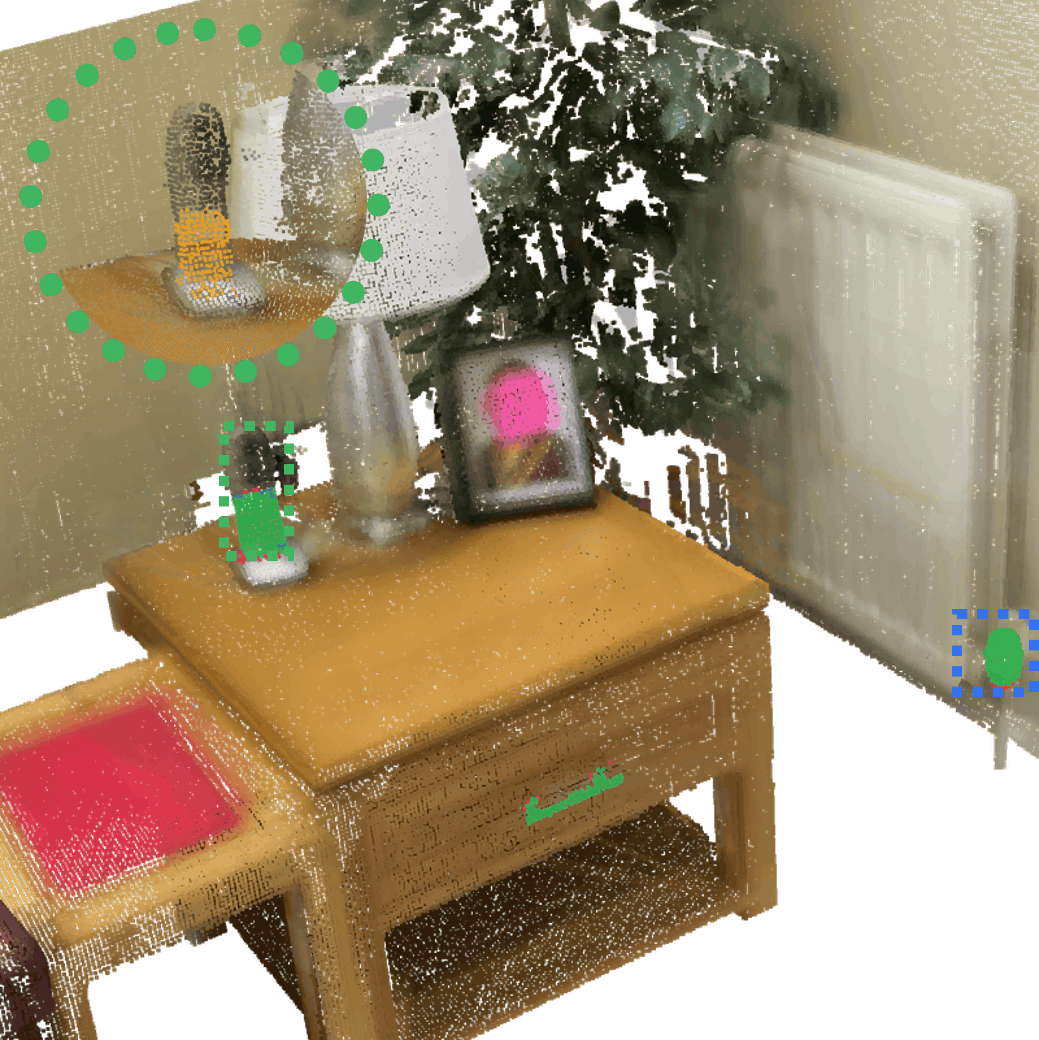} &
\includegraphics[width=0.19\textwidth,height=2.5cm]
    {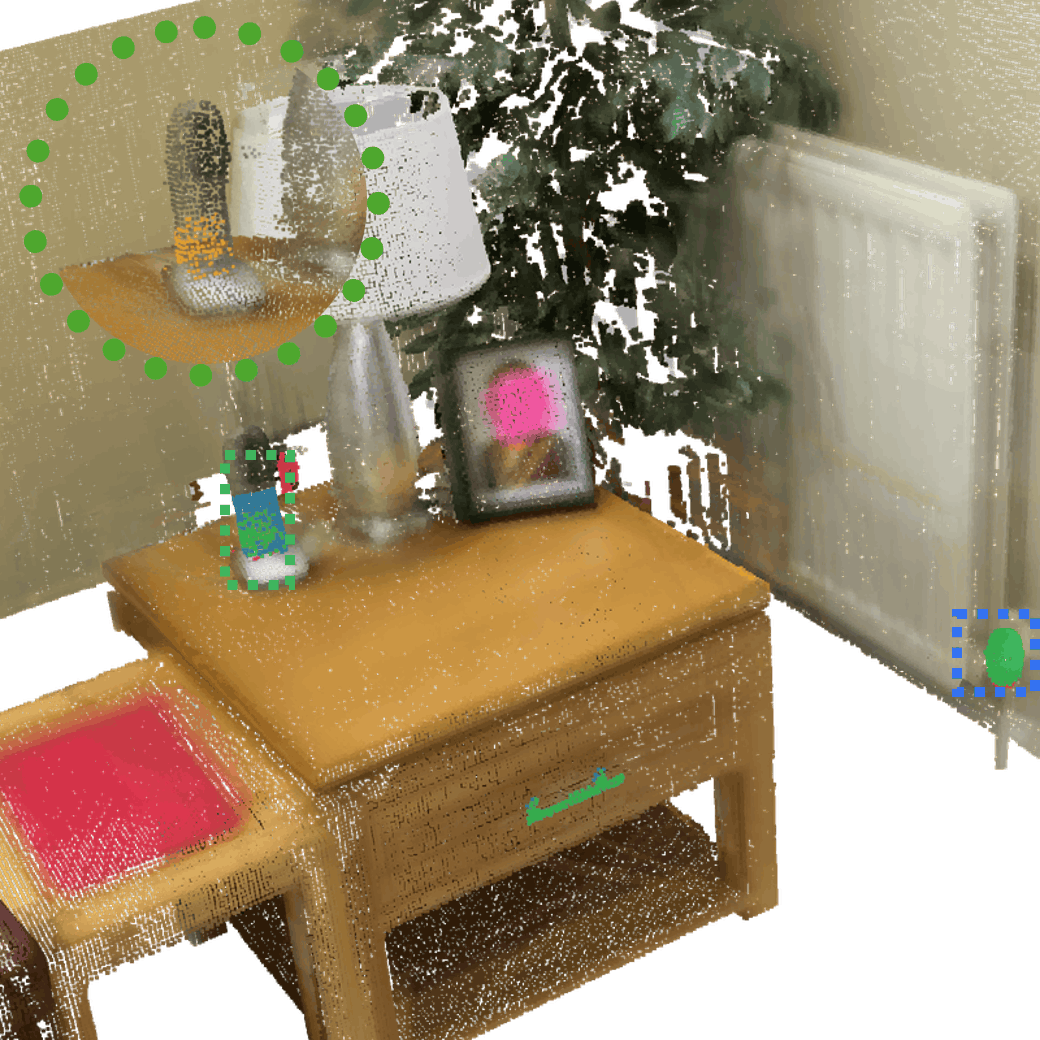} &
\includegraphics[width=0.19\textwidth,height=2.5cm]
    {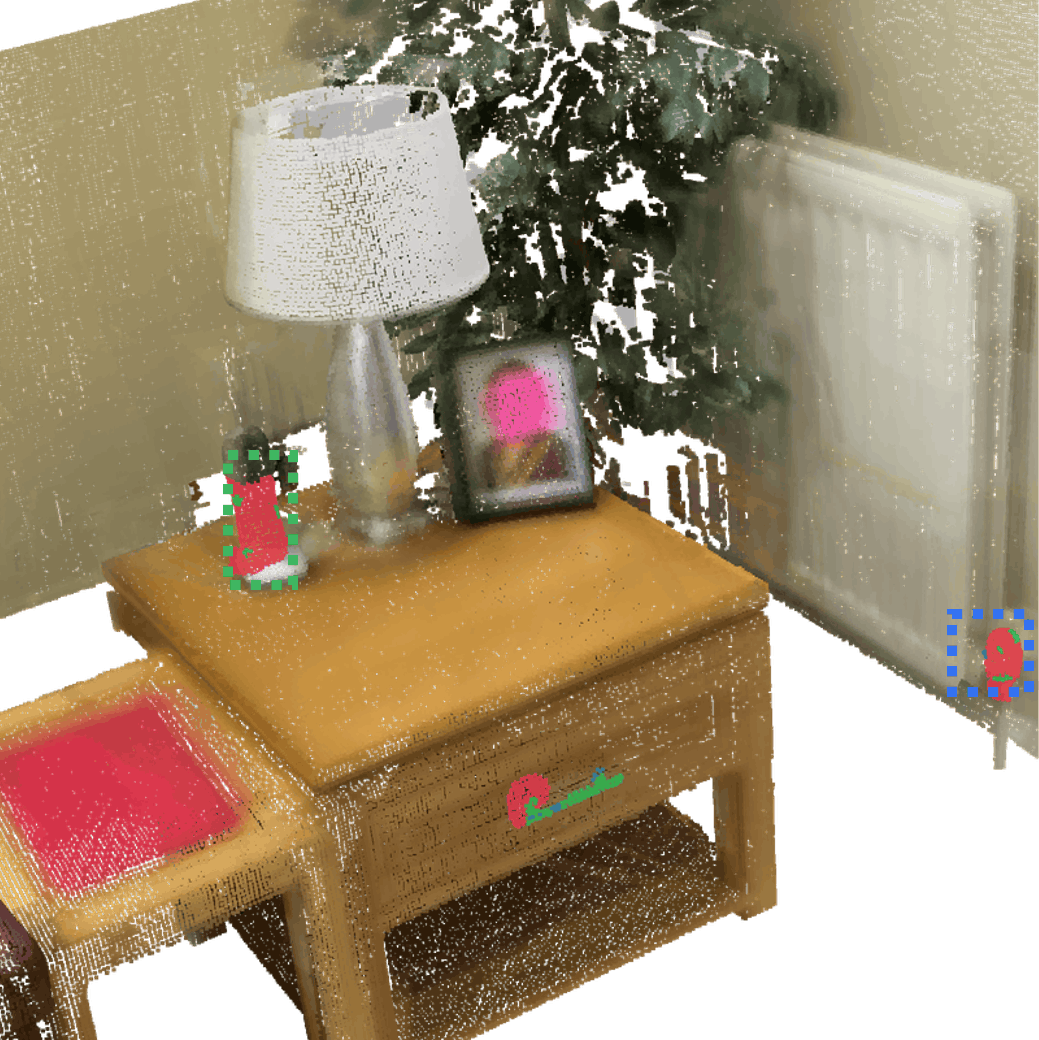} &
\includegraphics[width=0.19\textwidth,height=2.5cm]
    {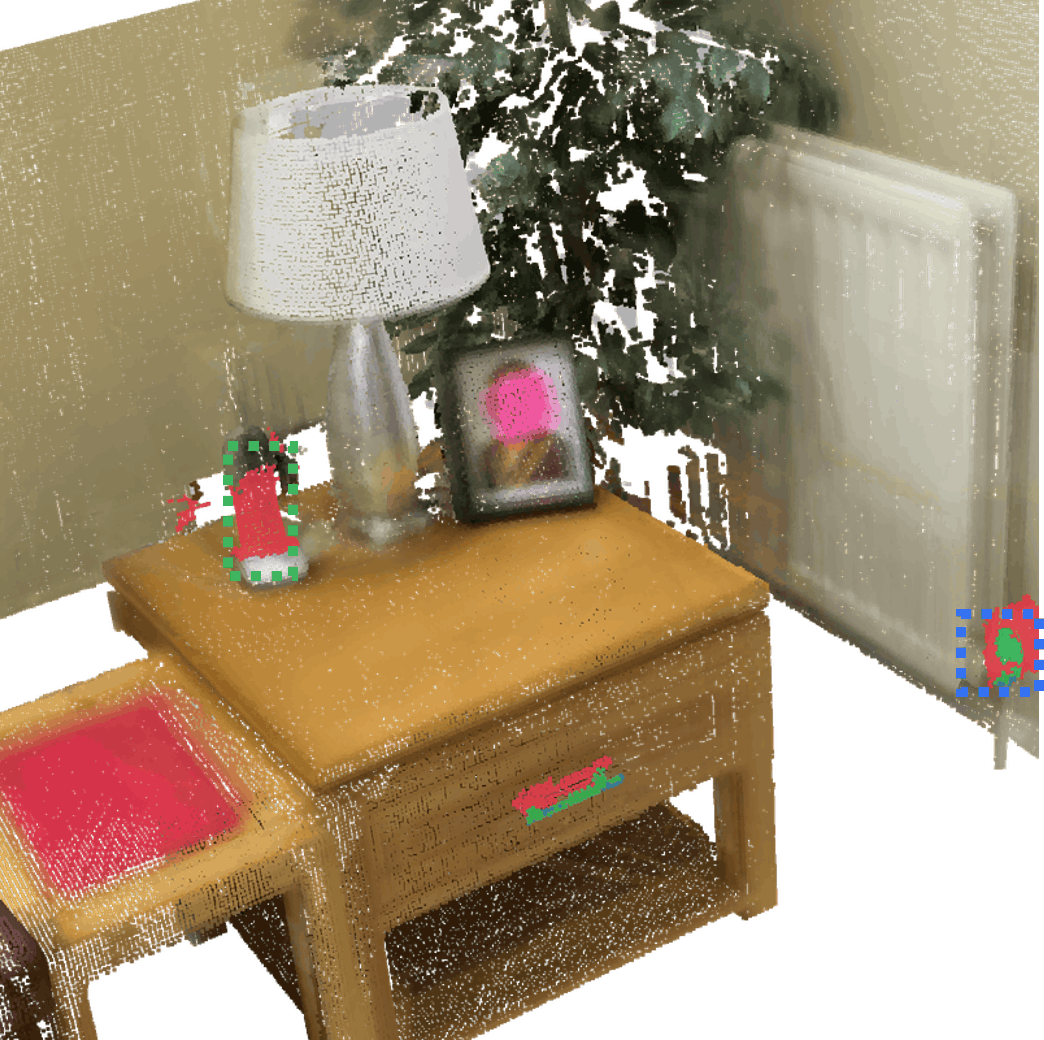}
\\[2pt]

{\small\textbf{Sample Frame with GT}} &
{\small\textbf{VGGT-$\forall^\dagger$}} &
{\small\textbf{Volt}} &
{\small\textbf{OVA-Fields}} &
{\small\textbf{LERF}}

\end{tabular*}
\vspace{-12pt}
\caption{
\textbf{Qualitative comparison on native point cloud.}
Columns compare image, VGGT-$\forall^\dagger$,
Volt~\cite{yilmaz2026volt},
OVA-Fields~\cite{su2025ovafields},
and LERF~\cite{kerr2023lerf}.
\textcolor{green!60!black}{Green},
\textcolor{red}{red}, and
\textcolor{blue}{blue} denote true positives,
false positives, and false negatives. Best viewed when zoomed.
}
\label{fig:qualitative_2}
\vspace{-18pt}
\end{figure*}

We evaluate VGGT-$\forall^\dagger$ on SceneFun3D's point clouds under the official Task 1 protocol ~\cite{scenefun3d_task1}.  We report results across all validation scenes and the hidden test set evaluated through the benchmark server ~\cite{scenefun3d_leaderboard}, comparing against other established methods ~\cite{delitzas2024scenefun3d, su2025ovafields, zhou2026opentrack3d, kerr2023lerf, yilmaz2026volt}. Performance is measured using class-averaged mAP and mAR. Qualitative samples are displayed at Fig.~\ref{fig:qualitative_2}.

\begin{table}[t]
\centering
\caption{Scan-assisted Volt inference cost. Point coverage counts unique raw scan points in aligned Volt patches.}
\vspace{-6pt}
\label{tab:volt_efficiency}
\resizebox{\columnwidth}{!}{%
\begin{tabular}{lrrrr}
\toprule
& \multicolumn{2}{c}{Ours}
& \multicolumn{2}{c}{Volt} \\
Attribute & Base & TTA & Base & TTA \\
\midrule
Point coverage, 30 visits (\%) & 18.74 & 18.74 & 100 & 100 \\
Total voxel exposures (B) & 0.02 & 3.27 & 0.54 & 7.43 \\
Max peak allocated VRAM (GiB) & 1.04 & 1.06 & 9.97 & 11.04 \\
Mean wall time/visit (s)  & 12.1 & 162.6 & 43.2 & 584.4 \\
\bottomrule
\end{tabular}}
\vspace{-12pt}
\end{table}

As shown in Table~\ref{tab:scenefun3d_val}, VGGT-$\forall^\dagger$ achieves 40.4 mAP, surpassing the full-scene Volt baseline (39.6), with consistent improvements across all six precision and recall metrics. Notably, our method improves $\mathrm{mAP}_{25}$ from 57.7 to 60.8 and $\mathrm{mAR}_{25}$ from 62.3 to 67.2, demonstrating stronger affordance instance coverage. This advantage extends to the hidden test set in Table~\ref{tab:scenefun3d_test}, where VGGT-$\forall^\dagger$ achieves 23.0 mAP, 34.5 $\mathrm{mAP}_{50}$, and 44.2 $\mathrm{mAP}_{25}$, outperforming Volt in overall AP and AP25 while matching its AP50.

To assess cross-dataset generalization without target-domain finetuning, we also evaluate on Articulate3D~\cite{halacheva2025articulate3d} using three shared affordance classes, where VGGT-$\forall^\dagger$  improves over Volt in both mAP and mAR, demonstrating consistent improvements across datasets.

Table~\ref{tab:volt_efficiency} further compares the computational efficiency. Guided by VGGT-$\forall$, our method covers only 18.74\% of native scan points across the validation scenes. This localized crop strategy reduces Volt-stage inference time from 584.4 to 162.6 seconds (3.6$\times$ speedup), and reduces peak VRAM from 11.04 GiB to 1.06 GiB. These results demonstrate that VGGT-$\forall^\dagger$ enables more selective native-point processing without sacrificing accuracy.


\subsection{Reconstructed Point-Cloud Evaluation}
\label{sec:exp_base}

\begin{figure*}[t]
\centering
\setlength{\tabcolsep}{0pt}

\begin{tabular}{
@{}c@{\hspace{0.005\textwidth}}
c@{\hspace{0.005\textwidth}}
c@{\hspace{0.020\textwidth}}
c@{\hspace{0.005\textwidth}}
c@{\hspace{0.005\textwidth}}c@{}
}


\begin{minipage}[c][0.20\textwidth][c]{0.12\textwidth}
\includegraphics[width=\linewidth]{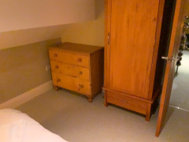}\par
\vspace{2pt}
\includegraphics[width=\linewidth]{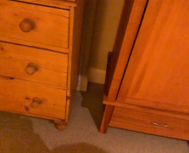}
\end{minipage}
&
\begin{minipage}[c][0.20\textwidth][c]{0.12\textwidth}
\includegraphics[width=\linewidth]{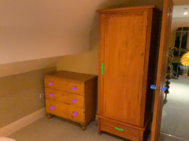}\par
\vspace{2pt}
\includegraphics[width=\linewidth]{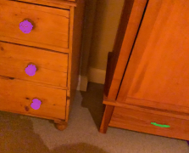}
\end{minipage}
&
\begin{minipage}[c][0.20\textwidth][c]{0.12\textwidth}
\includegraphics[width=\linewidth]{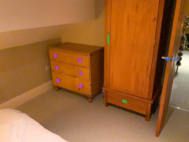}\par
\vspace{2pt}
\includegraphics[width=\linewidth]{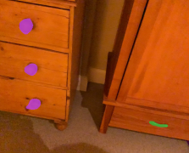}
\end{minipage}
&
\begin{minipage}[c][0.20\textwidth][c]{0.195\textwidth}
\centering
\includegraphics[
    width=\linewidth,
    height=\textwidth,
    keepaspectratio
]{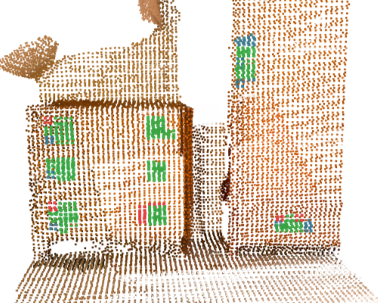}
\end{minipage}
&
\begin{minipage}[c][0.20\textwidth][c]{0.195\textwidth}
\centering
\includegraphics[
    width=\linewidth,
    height=\textwidth,
    keepaspectratio
]{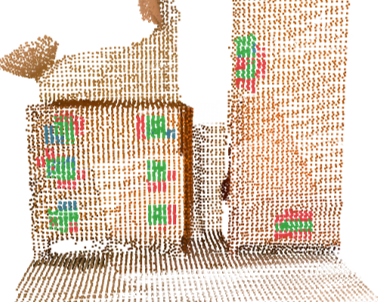}
\end{minipage}
&
\begin{minipage}[c][0.20\textwidth][c]{0.195\textwidth}
\centering
\includegraphics[
    width=\linewidth,
    height=\textwidth,
    keepaspectratio
]{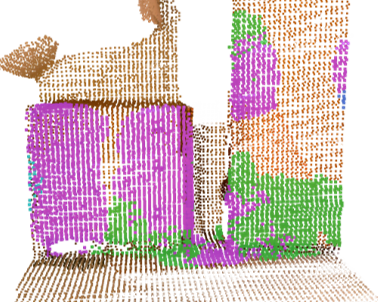}
\end{minipage}
\\[3pt]


\begin{minipage}[c][0.20\textwidth][c]{0.12\textwidth}
\includegraphics[width=\linewidth]{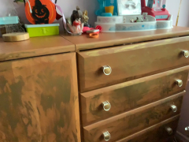}\par
\vspace{2pt}
\includegraphics[width=\linewidth]{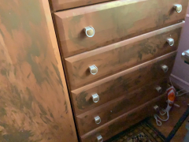}
\end{minipage}
&
\begin{minipage}[c][0.20\textwidth][c]{0.12\textwidth}
\includegraphics[width=\linewidth]{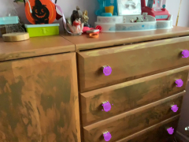}\par
\vspace{2pt}
\includegraphics[width=\linewidth]{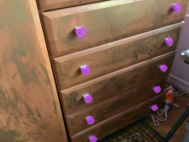}
\end{minipage}
&
\begin{minipage}[c][0.20\textwidth][c]{0.12\textwidth}
\includegraphics[width=\linewidth]{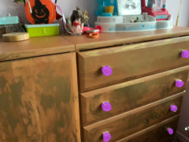}\par
\vspace{2pt}
\includegraphics[width=\linewidth]{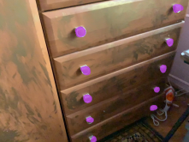}
\end{minipage}
&
\begin{minipage}[c][0.20\textwidth][c]{0.195\textwidth}
\centering
\includegraphics[
    width=\linewidth,
    height=\textwidth,
    keepaspectratio
]{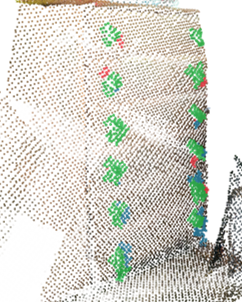}
\end{minipage}
&
\begin{minipage}[c][0.20\textwidth][c]{0.195\textwidth}
\centering
\includegraphics[
    width=\linewidth,
    height=\textwidth,
    keepaspectratio
]{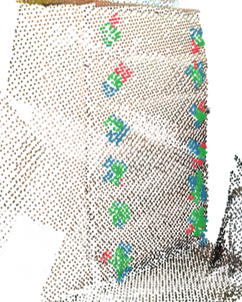}
\end{minipage}
&
\begin{minipage}[c][0.20\textwidth][c]{0.195\textwidth}
\centering
\includegraphics[
    width=\linewidth,
    height=\textwidth,
    keepaspectratio
]{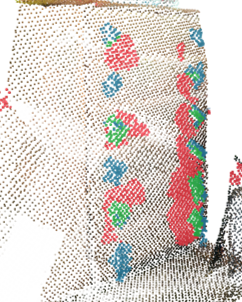}
\end{minipage}
\\[3pt]

\scriptsize RGB
&
\scriptsize Ground Truth
&
\scriptsize VGGT-$\forall$
&
\scriptsize VGGT-$\forall$
&
\scriptsize OVA-Fields
&
\scriptsize SegVGGT
\\

\end{tabular}

\vspace{-4pt}
\caption{
\textbf{Qualitative comparison of RGB-only 3D affordance discovery.}
For each scene, the left three columns show two RGB views,
their affordance annotations, and VGGT-$\forall$'s
predictions. The right three columns compare 3D affordance
instances predicted by VGGT-$\forall$,
OVA-Fields~\cite{su2025ovafields},
and SegVGGT~\cite{qu2026segvggt}
on reconstructed geometry. Best viewed when zoomed.
}
\label{fig:qualitative_base}
\vspace{-12pt}
\end{figure*}

We evaluate VGGT-$\forall$ on image-reconstructed SceneFun3D geometry, transferring ground-truth affordances from native scans through RGB-view correspondences. We compare against Mask3D-F~\cite{schult2023mask3d, delitzas2024scenefun3d} and Volt~\cite{yilmaz2026volt} fine-tuned on reconstructed geometry, OVA-Fields~\cite{su2025ovafields} and Fun3DU~\cite{corsetti2025fun3du} that leverage 2D affordance segmentation followed by 3D integration, and an affordance-adapted SegVGGT~\cite{qu2026segvggt}. Qualitative results are shown in Fig.~\ref{fig:qualitative_base}.

\begin{table}[t]
\centering
\caption{
SceneFun3D functionality segmentation on the validation set.
All methods are evaluated on reconstructed 3D point clouds.
}
\label{tab:scenefun3d_vggtval}
\resizebox{\columnwidth}{!}{
\fontsize{8}{10}\selectfont
\begin{tabular}{lccc}
\toprule
Method
& $\mathrm{mAP}$ $\uparrow$
& $\mathrm{mAP}_{50}$ $\uparrow$
& $\mathrm{mAP}_{25}$ $\uparrow$ \\
\midrule
Mask3D-F~\cite{delitzas2024scenefun3d, schult2023mask3d}
& 0.00 & 0.00 & 0.04 \\
Fun3DU~\cite{corsetti2025fun3du}
& \underline{2.82} & 6.46 & 19.72 \\
OVA-Fields~\cite{su2025ovafields}
& 2.15 & \underline{7.90} & \underline{54.00} \\
SegVGGT~\cite{qu2026segvggt}\   
& 0.19 & 1.22 & 16.84 \\
Volt~\cite{yilmaz2026volt}
& 0.01 & 0.04 & 0.60 \\
\midrule
\textbf{VGGT-$\forall$}
& \textbf{15.58} & \textbf{38.79} & \textbf{73.86} \\
\bottomrule
\end{tabular}
}
\vspace{-18pt}
\end{table}

As shown in Table~\ref{tab:scenefun3d_vggtval}, VGGT-$\forall$ achieves 15.58 mAP, 38.79 $\mathrm{mAP}{50}$, and 73.86 $\mathrm{mAP}{25}$, substantially outperforming existing approaches. Despite fine-tuning, Mask3D-F and Volt achieve near-zero mAP on reconstructed geometry, consistent with the degradation reported by SegVGGT~\cite{qu2026segvggt}, where OneFormer3D~\cite{kolodiazhnyi2024oneformer3d} drops from 30.2 to 2.5 mAP on VGGT-reconstructed ScanNet200~\cite{rozenberszki2022scannet200}. OVA-Fields and Fun3DU achieve only 7.90 and 6.46 mAP${50}$, respectively, while SegVGGT reaches 1.22 mAP${50}$ despite sharing a VGGT-based reconstruction paradigm. These results support decoding affordances directly from reconstruction representations with surface-aware supervision and cross-view consistency, rather than relying on scan-native segmentation or existing image-guided approaches. More comparisons~\cite{corsetti2025fun3du, wang2026afun, feng2026tfuns3d} are provided on the \textbf{Supplementary Material}.









\begin{table}[t]
\centering
\caption{
Ablation of VGGT-$\forall$ representations and
geometry-aware supervision under RGB-only inference.
}
\label{tab:representation_ablation}
\resizebox{\columnwidth}{!}{
\fontsize{7}{9}\selectfont
\begin{tabular}{l|ccc}
\toprule
Configuration
& mAP $\uparrow$
& $\mathrm{mAP}_{50}$ $\uparrow$
& $\mathrm{mAP}_{25}$ $\uparrow$ \\
\midrule
Spatial patches only
& 3.10 & 11.26 & 42.67 \\
+ Register tokens
& 7.07 & 20.78 & 67.72 \\
+ 512-resolution backbone
& 11.39 & 29.79 & 65.77 \\
+ Surface supervision
& 12.82 & 30.43 & 63.79 \\
+ Cross-view consistency
& 15.01 & 37.77 & 73.48 \\
\midrule
+ \textbf{Proposal Association (Final)}
& 15.58 & 38.79 & 73.86 \\
\bottomrule
\end{tabular}
}
\vspace{-18pt}
\end{table}

\subsection{Ablation Study}
\label{sec:exp_ablation}

Tables~\ref{tab:representation_ablation} and~\ref{tab:scan_ablation} examine the contributions of VGGT-$\forall$ and VGGT-$\forall^\dagger$. Using register tokens improves $\mathrm{mAP}$ from 3.10 to 7.07, displaying the benefit of scene-level features beyond patch tokens. Surface-level supervision, cross-view consistency, and proposal association further enhance affordance predictions, achieving 15.58 $\mathrm{mAP}$. For VGGT-$\forall^\dagger$, LoRA fine-tuning on Volt improves $\mathrm{mAP}_{25}$ from 47.67 to 58.17, highlighting the importance of adapting Volt to localized native-point crops. Rotation augmentation and balanced sampling further improve performance, reaching 54.37 $\mathrm{mAP}_{50}$. Together, these results demonstrate the complementary benefits of reconstruction-based affordance, geometric consistency, and native-point refinement.

\begin{table}[t]
\centering
\caption{
Ablation of affordance-guided native-point segmentation
on SceneFun3D validation scenes.
}
\label{tab:scan_ablation}
\resizebox{\columnwidth}{!}{
\fontsize{7}{8}\selectfont
\begin{tabular}{l|cc}
\toprule
Configuration
& $\mathrm{mAP}_{50}$ $\uparrow$
& $\mathrm{mAP}_{25}$ $\uparrow$ \\
\midrule
SpaceFormer Checkpoint & 21.83 & 26.94 \\
Point-SAM Checkpoint & 25.81 & 43.14 \\
Volt Checkpoint
& 44.89 & 47.67 \\
+ Finetuned Lora
& 50.80 & 58.17 \\
+ Rotation Augmentation
& 51.10 & 59.68 \\
\midrule
\textbf{Balanced Sampling (Final)}
& 54.37 & 60.76 \\
\bottomrule
\end{tabular}
}
\vspace{-18pt}
\end{table}

\subsection{Limitations and Future Works}
\label{exp:limitation}

VGGT-$\forall$ remains sensitive to occlusions, sparse viewpoints, and reconstruction inaccuracies, which may affect affordance localization. Additionally, VGGT-$\forall^\dagger$ depends on RGB-discovered candidates and may miss affordances outside these regions. Future work will focus on addressing uncertainty-aware geometric reasoning and robustness to sparse observations.
\section{Conclusion}
\label{sec:conclusion}

We introduced \textbf{VGGT-$\forall$}, a framework for joint 3D reconstruction and affordance understanding from uncalibrated RGB images, without requiring pre-existing 3D scans or task-specific queries. By developing spatial and semantic cues of VGGT-$\Omega$, our method enables scene-level affordance discovery directly from images. We further introduced \textbf{VGGT-$\forall^\dagger$}, an optional scan-assisted variant that transfers the results to native point clouds based on localized 3D segmentation. Experiments on SceneFun3D demonstrate substantial improvements over adapted baselines under our RGB-only evaluation protocol, while surpassing the strongest published baseline under SceneFun3D's official Task 1 benchmark. These results highlight the potential of image-based 3D representations for accessible scene-level affordance discovery, with refinement when detailed scans are available.

\section*{LLM Usage Statement}

OpenAI ChatGPT was used to assist with manuscript drafting, language refinement, and organization across the Introduction, Related Work, Methods, and Experiments sections. It also supported literature analysis, discussion of experimental comparisons, and LaTeX formatting of figures and tables. OpenAI Codex was also used to assist the authors in debugging codes and assisting visual rendering of qualitative examples. The authors independently conducted the research and training/inference experiments, verified the scientific content and reported results, and retain full responsibility for the manuscript.

{
    \small
    \bibliographystyle{ieeenat_fullname}
    \bibliography{main}
}


\end{document}